\documentclass[lettersize,journal]{IEEEtran}
\usepackage{amsmath,amsfonts}
\usepackage{algorithmic}
\usepackage{algorithm}
\usepackage{array}
\usepackage{textcomp}
\usepackage{stfloats}
\usepackage{url}
\usepackage{verbatim}
\usepackage{graphicx}
\usepackage{cite}

\usepackage{graphics}
\usepackage{subfigure}
\usepackage{booktabs}

\usepackage[small]{caption}

\usepackage{subcaption}
\usepackage{multirow}

\usepackage{tabularx}
\usepackage{ragged2e}

\usepackage{csquotes}

\usepackage{epsfig}

\usepackage{color}

\begin{document}

\title{Interruption-Aware Cooperative Perception for V2X Communication-Aided Autonomous Driving}

\author{
Shunli Ren, Zixing Lei, Zi Wang, Mehrdad Dianati ~\IEEEmembership{Senior Member}, Yafei Wang~\IEEEmembership{Member}, \\ Siheng Chen~\IEEEmembership{Senior Member}, Wenjun Zhang~\IEEEmembership{Fellow}

% \thanks{Manuscript received March 19, 2023.}
% \thanks{This work was supported in part by}% <-this % stops a space

\thanks{This research is partially supported by National Key R\&D Program of China under Grant 2021ZD0112801, NSFC under Grant 62171276 and the Science and Technology Commission of Shanghai Municipal under Grant 21511100900 and 22DZ2229005. \textit{(Corresponding author: Siheng Chen and Wenjun Zhang.)}}%
\thanks{Shunli Ren, Zixing Lei, Zi Wang, Siheng Chen and Wenjun Zhang are with the Cooperative Medianet Innovation Center, Shanghai Jiao Tong University, Shanghai, China (e-mail: renshunli@sjtu.edu.cn, chezacarss@sjtu.edu.cn, w4ngz1@sjtu.edu.cn, sihengc@sjtu.edu.cn, zhangwenjun@sjtu.edu.cn), and Siheng Chen is also with Shanghai AI Laboratory, Shanghai, China. }% <-this % stops a space
\thanks{Mehrdad Dianati is with the Warwick Manufacturing Group, The University of Warwick, Coventry CV4 7AL, U.K. (e-mail: M.Dianati@warwick.ac.uk).}
\thanks{Yafei Wang is with Shanghai Jiao Tong University, Shanghai, China (e-mail: wyfjlu@sjtu.edu.cn).}
% \thanks{Siheng Chen is with the Cooperative Medianet Innovation Center, Shanghai Jiao Tong University, Shanghai, China; and is also with Shanghai AI Laboratory, Shanghai, China (e-mail: sihengc@sjtu.edu.cn).}
% Mehrdad Dianati is with the Warwick Manufacturing Group, The University of Warwick, Coventry CV4 7AL, U.K. (e-mail: M.Dianati@warwick.ac.uk). Yafei Wang is with Shanghai Jiao Tong University, Shanghai, China (e-mail: wyfjlu@sjtu.edu.cn).

\thanks{This manuscript is an extension of a workshop paper\cite{iarcp} with no proceedings, which is not a formal publication.}
}



\maketitle

\begin{abstract}

Cooperative perception can significantly improve the perception performance of autonomous vehicles beyond the limited perception ability of individual vehicles by exchanging information with neighbor agents through V2X communication. However, most existing work assume ideal communication among agents, ignoring the significant and common \textit{interruption issues} caused by imperfect V2X communication, where cooperation agents can not receive cooperative messages successfully and thus fail to achieve cooperative perception, leading to safety risks. To fully reap the benefits of cooperative perception in practice, we propose V2X communication INterruption-aware COoperative Perception (V2X-INCOP), a cooperative perception system robust to communication interruption for V2X communication-aided autonomous driving, which leverages historical cooperation information to recover missing information due to the interruptions and alleviate the impact of the interruption issue. To achieve comprehensive recovery, we design a communication-adaptive multi-scale spatial-temporal prediction model to extract multi-scale spatial-temporal features based on V2X communication conditions and capture the most significant information for the prediction of the missing information. To further improve recovery performance, we adopt a knowledge distillation framework to give explicit and direct supervision to the prediction model and a curriculum learning strategy to stabilize the training of the model. Experiments on three public cooperative perception datasets demonstrate that the proposed method is effective in alleviating the impacts of communication interruption on cooperative perception. V2X-INCOP outperforms state-of-the-art cooperative perception methods and has a cooperative perception gain up to 14.06\%, 13.9\%, and 12.07\% over individual perception on average of different packet drop rates on OPV2V, V2X-Sim, and Dair-V2X datasets, respectively.

\end{abstract}


\section{Introduction}

\begin{figure*}[t]
	\centering
	\subfigure[Interruption issue.]{
		\begin{minipage}[b]{0.43\textwidth}
			\includegraphics[width=1\textwidth]{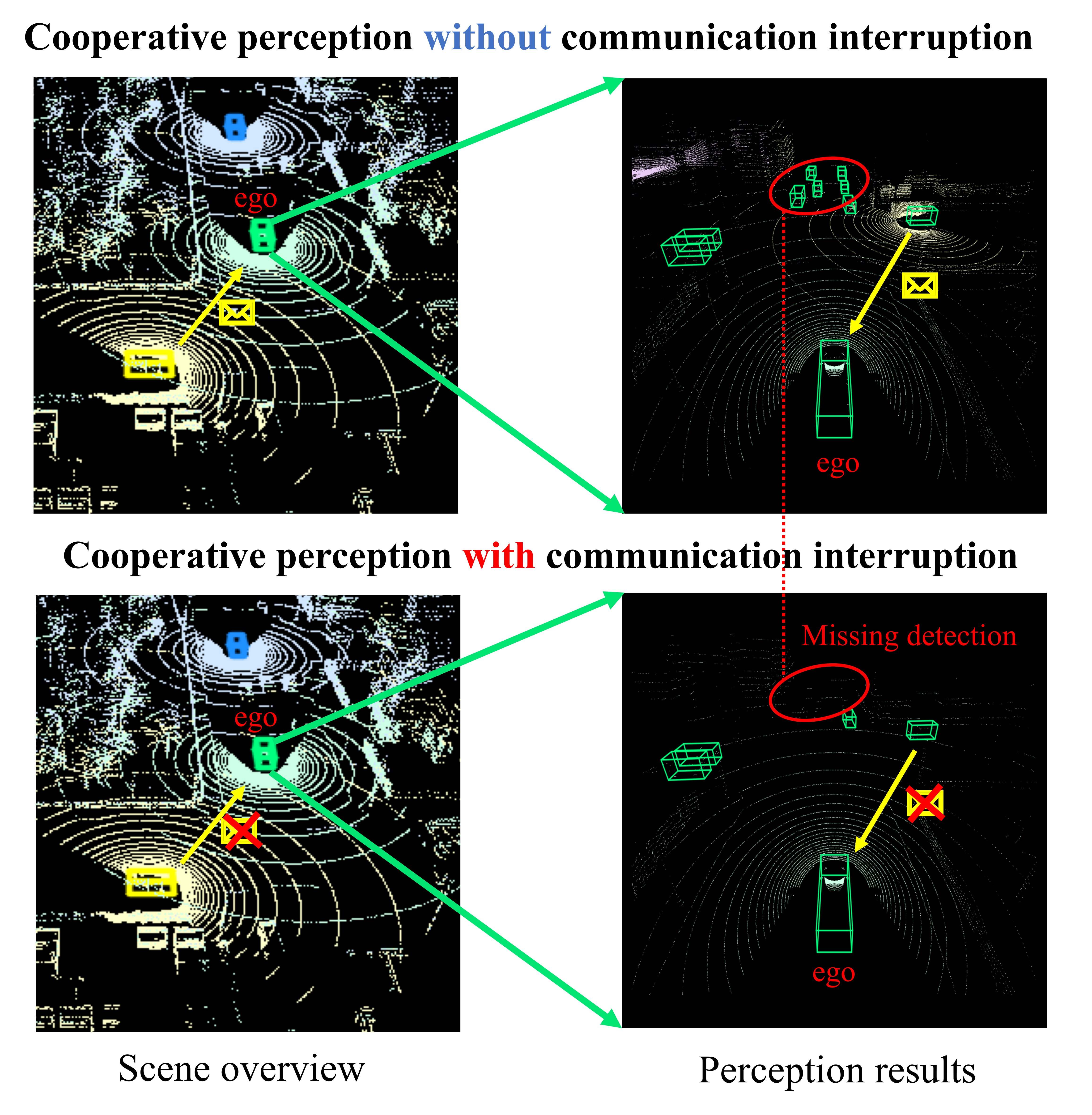}
		\end{minipage}
		% \label{fig:issue}
	}
    	\subfigure[Adverse effect of communication interruption.]{
    		\begin{minipage}[b]{0.51\textwidth}
   		 	\includegraphics[width=1\textwidth]{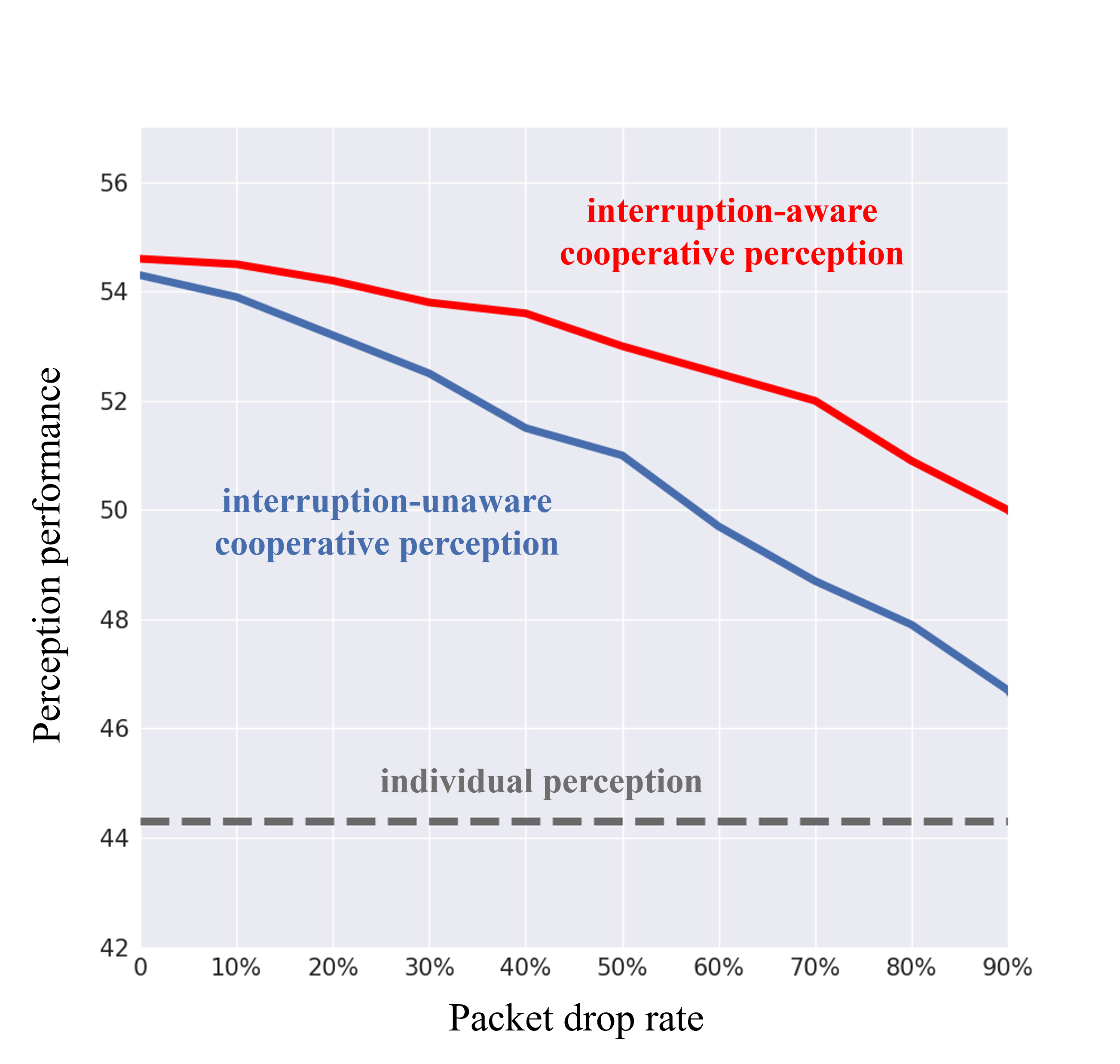}
    		\end{minipage}
		% \label{fig:performance degrade}
    	}
    \caption{
(a) \textbf{Communication interruption issue.} With successful communication, the perception range of the green vehicle is expanded with the help of the supportive message from the yellow vehicle, effectively addressing occlusion and long-range challenges. When communication interruption between the two vehicles happens, the green vehicle fails to receive the cooperative message, causing missing detections 
 and security risks.
(b) \textbf{Cooperative perception performance with different packet drop rates on the V2X-Sim1.0~\cite{li2021learning} dataset.} It is evident that the performance experiences significant degradation as a direct consequence of communication interruptions, represented by the blue curve. Fortuitously, the proposed interruption-aware cooperative perception system, represented by the red curve, adeptly mitigates the observed degradation.
     }
     \label{fig:interruption issue}
\end{figure*}

Autonomous Vehicles (AVs) and Automated Driving Systems (ADSs) promise to improve the safety, efficiency, and accessibility of future transport systems. AVs and ADSs rely on accurate and robust understanding of their surroundings, i.e., perception, for safe, deficient, and comfortable motion planning~\cite{5959985,10093116,10197250}. 
Poor perception can result in traffic accidents, threatening human safety~\cite{tesla,highway}, and other forms of damage. While rapid developments of advanced sensors and perception algorithms~\cite{8621614,9307334,10038646} resulted in remarkable progress in AV/ADS technology development in recent years, the inadequacy of the existing solutions remains to be a major obstacle on the road to safe employment of technology on public roads.

One inherent limitation is associated with the perception ability of a single vehicle being fundamentally limited. AVs/ADSs that relay on sole ego-vehicle-based perception suffer from several perception-related issues, such as: 1) \textit{long-range issue}: objects at long distances are harder to perceive due to fewer measurements. For example, objects at a long distance may be covered by only a few LiDAR points or image pixels and even uncovered; 2) \textit{occlusion issue}: some objects may be occluded by obstacles in line-of-sight in crowded or complex environments; 3) \textit{sensor malfunction and noise issue}: perception sensor may temporally break down due to internal and external factors, such as circuit failure and lens smudge.  Also, the field of view of the onboard sensors may be impaired by lighting conditions and other sensor noises. These issues may result in missed or incorrect perception, and thus cause a cascade failure of the downstream modules of autonomous driving, such as tracking, prediction, planning and decision-making.

Cooperative perception is a promising remedy to address the above fundamental issues of individual perception~\cite{ren2022collaborative}. 
With Vehicle-to-everything (V2X) communication, connected and autonomous vehicles (CAVs) and roadside units (RSUs) share perception information, and cooperate to achieve a highly accurate and robust perception beyond the individual perception range and line of sight.
It has been demonstrated that the application of cooperative perception to V2X communication-aided autonomous driving can increase perception accuracy and transportation safety~\cite{shan2020demonstrations,schiegg2020collective,10091215}. 
Recently, cooperative perception has attracted significant attention from the research community, AV technology development companies, and road operators. 
Some recent studies proposed cost-effective and efficient cooperation strategies to achieve good perception performance for AVs~\cite{wang2020v2vnet,li2021learning,Where2comm22} while some other works leveraged cooperative perception to improve connected and automated vehicles systems and vehicle-roadside cooperation technologies, prompting better autonomous driving~\cite{9735412,9497693,fang2024fedrsu}.

Successful deployment of existing cooperative perception methods mostly depends on ideal V2X communication conditions assumption.
However, real-world communication seldom attains a state of perfection. Even though V2X communication technology is developing rapidly~\cite{kose2021beam,9611166}, V2X link failures and communication errors are still inevitable and~\emph{random temporary link interruptions} are common communication problems~\cite{9310690,9514510}. 
Some previous works~\cite{9385955,9447840,9068410} have studied some of the common V2X communication failures in various scenarios, such as: 1) when too many vehicles exchange messages in a certain area, the communication channels become crowded and packet drops can happen; 2) packet drops are likely to occur when  V2X communication range is relatively large; 3) when V2X communication is attacked by malicious users, valid users can not receive accurate messages in time. In such scenarios, vehicles may fail to receive messages from others in time, resulting in the \textbf{communication interruption issue}.
In this work, we formulate the communication interruption issue as that each communication link connecting any two cooperating nodes could possibly be interrupted, resulting in the nodes being unable to receive messages from all of their neighboring nodes.
Hence, the performance of cooperative perception may be severely degraded, and the downstream tasks such as tracking and trajectory prediction may be affected by V2X link failures, causing a cascading failure in the system and threatening the safety and efficiency of the autonomous driving system. Figure~\ref{fig:interruption issue} provides an example of the communication interruption issue and its adverse effect.
However, the communication interruption issue in cooperative perception has been not well studied and resolved to the best of our knowledge, hindering the practical application of cooperative perception.

To fill this gap, this work formulates the problem of cooperative perception with random communication interruptions and proposes V2X communication INterruption-aware COoperative Perception (V2X-INCOP), a novel cooperative perception system robust to V2X communication interruption for V2X communication-aided autonomous driving, facilitating the application of cooperative perception in real-world scenarios.
It's worth emphasizing that the specific goal is to mitigate the adverse effects arising from V2X communication interruptions from the perspective of cooperative perception algorithm design, instead of a communication perspective. The proposed V2X-INCOP leverages historical information to recover the missing information due to the communication interruption. The underlying motivation is that we can leverage relevant information from the cooperation history to infer the currently missing information. In the proposed V2X-INCOP system, a communication-adaptive prediction model is specifically designed to achieve reliable recovery by extracting multi-scale spatial-temporal features from historical information. To improve the performance of the prediction model, we leverage knowledge distillation to provide explicit supervision to the model, guiding the model to estimate the current state of the historical information. We employ a curriculum learning strategy to help the model learn more efficiently and converge rapidly. The training process commences with a low probability of interruption, gradually escalating the range of interruption probabilities.

In the field of intelligent transportation and autonomous driving, LiDAR plays an important role in environmental perception and situational awareness because it can collect high-resolution 360-degree data in the 3D environment, which has been widely applied in the real world. To establish a cooperative perception system with practical applicability, we focus on the task of LiDAR-based 3D object detection in this work since it is an important task in the perception of autonomous driving.
To validate the proposed method, we conduct experiments on three public large-scale cooperative perception datasets, V2X-Sim~\cite{li2022v2x}, OPV2V~\cite{xu2022opencood}, and Dair-V2X~\cite{yu2022dair}. The three datasets provide autonomous driving scenarios where multiple vehicles and roadside units can collect perception data simultaneously, which enables multi-agent cooperation scenarios. Each dataset includes more than 5K frames of LiDAR point clouds and supports the task of LiDAR-based 3D object detection. V2X-Sim and OPV2V are both simulated datasets generated by CARLA~\cite{dosovitskiy2017carla}. V2X-Sim includes V2V and V2I scenarios in crowded downtown and suburb highways and OPV2V includes V2V scenarios with diverse city traffic configurations. Dair-V2X collects data from real-world vehicle-infrastructure cooperation scenarios. The experimental results show that the proposed V2X-INCOP system outperforms existing cooperation strategies at various packet drop rates(PDRs) and effectively alleviates the effect of V2X communication disruptions. On average of different PDRs, V2X-INCOP has a cooperative perception gain of up to 14.06\% over individual perception.

The contributions of this paper can be summarized as follows:

1) We propose V2X-INCOP, a cooperative perception system robust to communication interruption for V2X communication-aided autonomous driving. To our best knowledge, this is the first work to address the communication interruption issue in cooperative perception, which promotes the practical application of cooperative perception.

2)  The proposed V2X-INCOP leverages a novel communication adaptive spatial-temporal prediction model to recover the missing information from historical information. To further improve the performance, knowledge distillation is employed to provide the ground truth to the prediction model, and a curriculum learning strategy is adopted to achieve better generalization to different communication interruption conditions.

3) We conduct extensive experiments on three large-scale cooperative perception datasets and the results show that the proposed V2X-INCOP brings significant benefits to mitigate the adverse effects of communication interruption, {which outperforms the existing state-of-the-art cooperative perception methods in the presence of communication interruption, and has a cooperative perception gain up to 14.06\%, 13.9\%, and 12.07\% over individual perception on average of different PDRs on OPV2V, V2X-Sim, and Dair-V2X datasets, respectively.

The rest of this paper is organized as follows. Section~\ref{sec:related} discusses related works. Then, Section~\ref{sec:problem} gives the system model and problem formulation of cooperative perception with communication interruption. Section~\ref{sec:method} provides an overview of the proposed system as well as details of each component design. Section~\ref{sec:exp} presents the experiment results and discussion which validates the effectiveness of V2X-INCOP. Finally, Section~\ref{sec:conclude} concludes the main contributions and points out directions for further research.

\section{Related Works}\label{sec:related}

This section provides an overview of the existing related works in the literature. The section is divided into three subsections. Subsection~\ref{subsec:cp} introduces existing cooperative perception strategies, Subsection~\ref{subsec:com} introduces  the development of V2X communication and Subsection~\ref{subsec:3dod} introduces the development of Lidar-based 3D object detection algorithms.

\subsection{Cooperative perception}\label{subsec:cp}

Cooperative perception can effectively improve the perception performance beyond the limitation of the individual agent in multi-agent systems. 
The relevant work on cooperative perception primarily focuses on four aspects: exploring cooperation modes, prompting efficiency and performance, creating datasets, and enhancing robustness and security.

The existing cooperative perception methods can be divided into four cooperation modes according to what messages are shared: early cooperation, intermediate cooperation, late cooperation, and mixed cooperation.
Cooper~\cite{chen2019cooper} firstly proposes an early cooperation strategy, and F-Cooper~\cite{chen2019f} improves it with an intermediate cooperation strategy. Arnold et al.~\cite{arnold2021fast} propose a fast and robust registration method for cooperation on the point-clouds level.
To prompt the efficiency and performance of cooperative perception, some work design more efficient cooperation strategies for better trade-offs between communication cost and perception performance.
V2VNet~\cite{wang2020v2vnet} leverages a vanilla spatial-aware graph neural network to aggregate cooperation messages, and DiscoNet~\cite{li2021learning} proposes a cooperation graph with matrix-valued edge weight to model the importance of different spatial cells and improve perception performance with knowledge distillation. FPV-RCNN~\cite{yuan2022keypoints} and Where2comm~\cite{Where2comm22} select important features to save communication costs based on key points and spatial confidence map, respectively. Who2com~\cite{liu2020who2com} and When2com~\cite{liu2020when2com} construct communication groups with attention mechanism, and  COOPERNAUT~\cite{cui2022coopernaut} and V2X-Vit~\cite{xu2022v2xvit} conduct information fusion with transformer. 
To validate the performance of cooperative perception more legitimately, some large-scale cooperative perception datasets have been released, including simulation datasets (V2X-Sim~\cite{li2022v2x}, OPV2V~\cite{xu2022opencood}) and real-world dataset (DAIR-V2X~\cite{yu2022dair}). 
To ensure the robustness and safety of cooperative perception, some works consider some practical constraints in the real world. Vadivelu et al.~\cite{vadivelu2021learning} and Lu et al.~\cite{lu2022robust} study robust cooperative perception in the presence of pose errors.  Li et al.~\cite{10077757} study cooperative perception with lossy messages. SyncNet~\cite{lei2022latency} and CoBEVFlow~\cite{wei2024asynchrony} propose cooperative perception with time delay and asynchronous cooperation messages. ROBOSAC~\cite{Li2023AmongUA} and MADE~\cite{zhao2023malicious} propose defense strategies to attackers and malicious agents in the cooperative perception systems. HEAL~\cite{lu2024extensible} establishes a extensible cooperative perception framework for heterogeneous agent types.

However, there is no previous work studying cooperative perception with communication interruption, which is a fundamental and common communication issue in the real world and could adversely impact the performance of cooperative perception. To fill this gap, we propose an interruption-aware robust cooperative perception (V2X-INCOP) system for V2X communication-aided autonomous driving, which can alleviate the impact caused by communication interruption issues, by recovering the missing information due to communication interruptions. In a previous version~\cite{iarcp}, we propose to recover the missing information with three loosely structured modules, completion model, prediction model, and mask constraint. In this work, the proposed V2X-INCOP obtains communication-adaptive historical information by an attentive fusion model and captures comprehensive spatial-temporal features by a multi-scale spatial-temporal prediction model, achieving better performance. Moreover, this version provides a more profound analysis of the motivation and significance of cooperative perception in the presence of communication interruption, offered a more comprehensive exposition of our proposed approach and adds a substantial amount of comprehensive experimental results with detailed analysis and discussion.

\subsection{V2X Communication}\label{subsec:com} 
V2X communications are the communication basis of the application of cooperative perception to intelligent transportation systems, which supports the exchanges of messages between vehicles and their neighbor agents such as vehicles(V2V), infrastructures(V2I), and pedestrians(V2P)~\cite{9645261,9709116}. There are two kinds of mainstream V2X communication technologies: DSRC-based communications and cellular-based communications. Dedicated short-range communications(DSRC)~\cite{kenney2011dedicated} technology conducts short information exchange among devices equipped with 802.11p chips and provides V2X communication service. Cellular communication such as LTE and 5G can provide communication with high mobility and a large number of vehicles in a cell. 
While V2X communication technology continues to evolve,  some issues remain due to fundamental limitations. DSRC can only work well in a short communication range so communication may face interruptions when the distance between two vehicles is considerable. Centralized mode of cellular communications may raise overload~\cite{9217500}. The elevated load on the radio channel is likely to result in packet drop and substantial communication latency, which can degrade the accuracy and timeliness of cooperative perception and pose serious safety threats.

It is evident that communication interruption issues are bound to persist over an extended period, especially in the case of long-range, high-load, and complex environments, which will disrupt cooperative perception and pose potential safety hazards. However, there is no previous work on cooperative perception considering this point. To fill this gap, this work focuses on ameliorating the impact of communication interruption through the design of cooperative perception algorithms, with which the performance of cooperative perception can keep robust to interruption, instead of avoiding interruption from a communication perspective.

\subsection{LiDAR-based 3D object detection} \label{subsec:3dod}

In this work, we focus on the task of LiDAR-based 3D object detection, because it is an important and challenging task for the perception module of autonomous driving. LiDAR point clouds can capture the 3D structures of the scenes for a better understanding of the surrounding environment. The difficulty of LiDAR-based 3D object detection is how to extract semantic features from irregular LiDAR point clouds. Related works can be divided into two groups according to point cloud representation methods, \textit{point-based} methods and \textit{grid-based} methods. \textit{Point-based} methods, like PointNet~\cite{qi2017pointnet}, involve the direct processing of input point clouds to generate a sparse representation and obtain a feature vector for each point through the aggregation of features from its neighboring points.  PointNet++~\cite{qi2017pointnet++} is a multi-scale PointNet by learning local and global features with different contextual scales. PointRCNN~\cite{Shi2019PointRCNN3O} learns discriminative point-wise features with PointNet++ and conducts object detection with the following proposal generation and box refinement.
\textit{Grid-based} methods, like VoxelNet, divide the point clouds into equally spaced 3D grids and extract features of a group of points within each grid. Then the features can be extracted by convolution neural networks.
Pointpillars~\cite{lang2019pointpillars} engages in learning the representation of point clouds within a \enquote{pillar}, denoting a vertical column, instead of voxels. In this way, a point cloud can be converted to a sparse pseudo image and thus saving many computation resources.
PV-RCNN~\cite{shi2020pv} uses a 3D CNN to extract 3D features into a small set of key points and abstract proposal-specific features from the key points, which takes advantage of both kinds of methods.
Our proposed method mainly focuses on the task of LiDAR-based 3D object detection. Most existing methods focus on improving individual detection performance.
In this work, we combine the backbone networks of state-of-the-art lidar-based 3D object detection methods with our interruption-robust cooperative perception resolution to build the proposed interruption-aware cooperative perception system.

\newcolumntype{L}{>{\RaggedRight\hangafter=1\hangindent=0em}X}

\begin{table}

\caption{Explanation of notations and the abbreviations for the proper nouns.}
\label{tab:not}
\begin{tabularx}{0.48\textwidth}{ p{2cm} L }
\toprule
Notation  & Definition \\
\midrule
  $a_i$ & Cooperation node (vehicle or roadside unit) \\
  $N$ & Number of cooperation nodes in the scene \\
  $t$ & Timestep \\
  $\mathcal{N}_{i}$ & Set of neighbor cooperation nodes of node $a_i$ \\
  $\mathcal{R}^{(t)}_i$   & Set of nodes from which node $a_i$ can successfully receive messages at timestep $t$ \\

  $\mathbf{X}_{i}^{(t)}$   & Raw sensor data observed by node $a_i$ at timestep $t$ \\
  $\mathbf{F}_{i}^{(t)}$   & Messages transmitted from node $a_i$ at timestep $t$, \textit{i.e.}, features of $\mathbf{X}_{i}^{(t)}$  \\
  $\mathbf{H}_{i}^{(t)}$   & Features after information fusion \\
  $\widehat{\mathbf{Y}}_{i}^{(t)}$   & Perception results of node $a_i$ at timestep $t$ \\
  $\mathbf{Y}_{i}^{(t)}$   & Ground truth of $\widehat{\mathbf{Y}}_{i}^{(t)}$ \\
  \midrule
  Abbreviation   & Full name \\
  \midrule
  V2X & Vehicle-to-everything \\
  V2V & Vehicle-to-vehicle \\
  V2I & Vehicle-to-infrastructure \\
  V2X-INCOP &  V2X communication INterruption-aware COoperative Perception \\
  PDR &  packet drop rate \\
  SAFF & Spatial Attentive Feature Fusion \\
  KD & Knowledge distillation \\
\bottomrule
\end{tabularx}

\end{table}

\section{System Model and Problem Formulation}\label{sec:problem} 

This section introduces the foundational elements, the basic workflow and the goals of the cooperative perception system model and gives a problem formulation of cooperative perception with communication interruption issues, a not well-formulated problem in previous work, which provides the preliminary information of cooperative perception and clearly defines the problem that this work aims to address.
The \enquote{Pipeline of Cooperative Perception} located in the lower part of Figure~\ref{fig:overview} can be referred to for a better understanding of the information flow. A notation table is given in Table~\ref{tab:not} to explain the meanings of the notations and the abbreviations for the proper nouns.

We consider there are $N$ cooperation nodes, including vehicles and roadside units, in the cooperative perception system. Each node, denoted as $a_i$, maintains a set of neighboring nodes $\mathcal{N}_{i}$. At each timestep $t$, each node $a_i$ uses an encoder $f_{\mathrm{encode}}$ to exact a set of desired features $\mathbf{F}_{i}^{(t)}$ from its local observation of surrounding environment $\mathbf{X}_{i}^{(t)}$, defined as,
\begin{equation*}
    \mathbf{F}_{i}^{(t)} = f_{\mathrm{encode}}(\mathbf{X}_{i}^{(t)}).
\end{equation*}

Once generated, these features are then transmitted to neighboring nodes. 

With ideal communication, each node can successfully receive messages from all its neighbors at each timestep.
Subsequently, each node $a_i$ fuses the received features and its own features using a fusion model denoted as $f_{\mathrm{fuse}}$ to generate a fused feature $H_{i}^{(t)}$, represented as:
\begin{equation*}
    \label{eq:fusion_ideal}
    \mathbf{H}_{i}^{(t)} = f_{\mathrm{fuse}}(\{ \mathbf{F}_{j}^{(t)}\}_{j \in \{i\} \cup \mathcal{N}_i}).
\end{equation*}

\begin{figure}
\centering
\includegraphics[width=\linewidth]{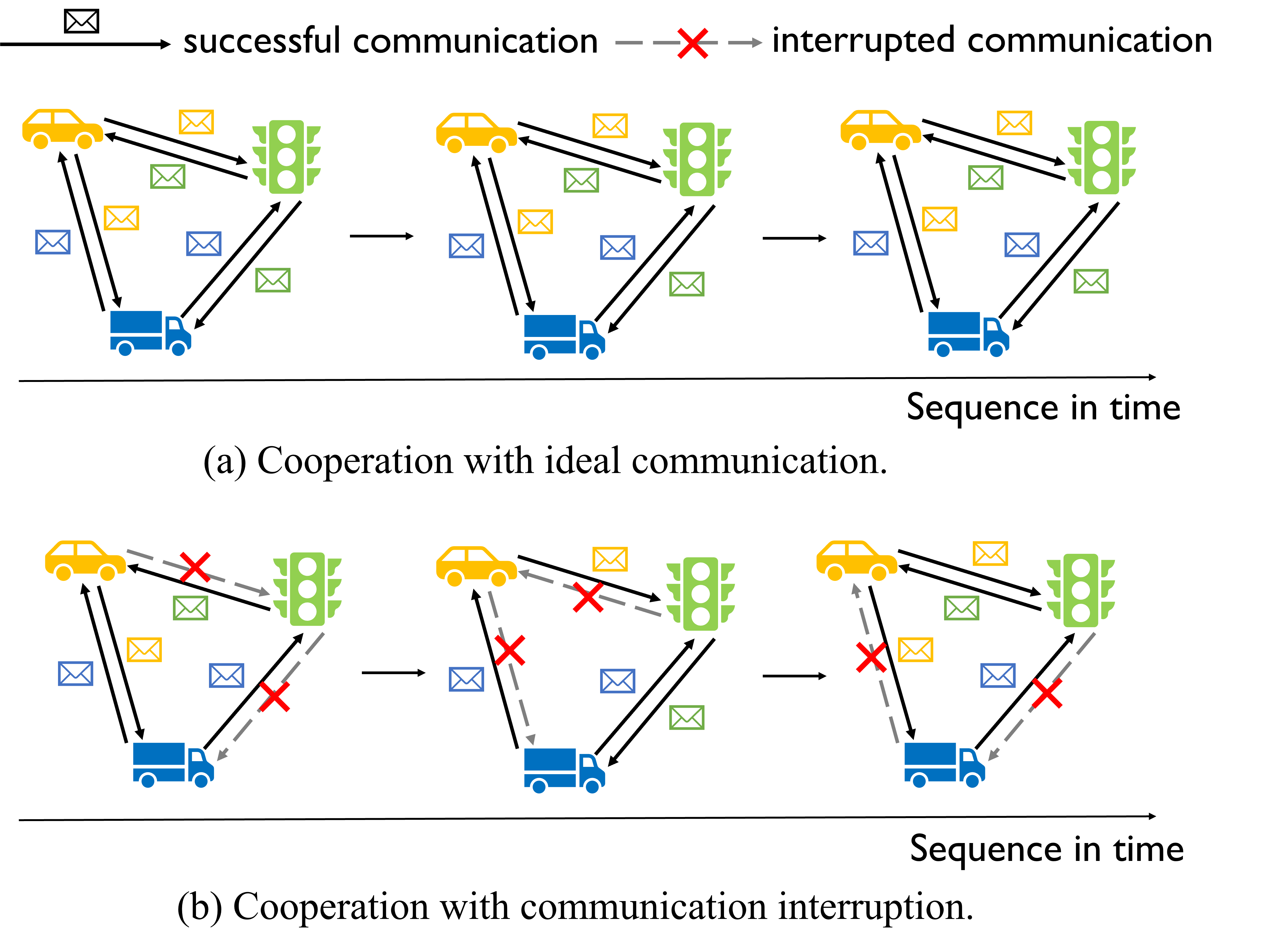}
\caption{\textbf{Illustration of cooperative perception with/without communication interruption.} (a) With ideal communication, each cooperation node can succeed to receive messages from all its neighbours at each timestep. (b) With consideration of communication interruption, interruption randomly happens so that each cooperation node can only receive messages from parts of its neighbours. Communication between each agents pair at each step interrupts with a certain interruption probability.}
\label{fig:graph}%
\end{figure} 

When considering communication interruption, we first formulate the communication interruption we study in this work. The communication interruption between two cooperation nodes is defined as, at a certain timestep, one node is unable to receive messages sent by another node.

With the occurrence of communication interruptions, each communication link connecting any two cooperating nodes could possibly be interrupted at each timestep, resulting in the nodes being unable to receive messages from their neighboring nodes, as depicted in Figure~\ref{fig:graph}. 
We formulate the severity of communication interruption with a packet drop rate $0 < p < 1$, which is also an interruption probability indicating how likely communication interruption will occur between each two cooperation nodes at each timestep. The packet drop means a message from a node fails to be received by its target node, which indicates a communication interruption. Then, we can obtain the set of cooperation nodes from which node $a_i$ can successfully receive messages at timestep $t$, $\mathcal{R}^{(t)}_i \in \mathcal{N}_i$, by uniformly sampling nodes from $\mathcal{N}_i$ with the probability of $1 - p$.
In this case, node $a_i$ can receive messages from nodes $a_j \in \mathcal{R}_{i}^{t}$, and fuse messages from other nodes with its own perception features to produce the fused feature $\mathbf{H}_{i}^{(t)}$. 
The fusion process can be formulated as follows:
\begin{equation*}
    \label{eq:fusion_interruption}
    \mathbf{H}_{i}^{(t)} = f_{\mathrm{fuse}}(\{ \mathbf{F}_{j}^{(t)}\}_{j \in \{i\} \cup \mathcal{R}^{(t)}_i}).
\end{equation*}

Lastly, a decoder denoted as $f_{\mathrm{decode}}$ is employed to decode the fused features, resulting in the perception results $\widehat{\mathbf{Y}}_{i}^{(t)}$ for each node, expressed as follows,
\begin{equation*}
    \widehat{\mathbf{Y}}_{i}^{(t)} = f_{\mathrm{decode}}(\mathbf{H}_{i}^{(t)}).
\end{equation*}

The ground truth of the perception results for node $a_i$ at time step $t$ is denoted as $\mathbf{Y}_{i}^{(t)}$.

With random communication interruption, cooperative perception performance will be significantly degraded because of inadequate message passing. Therefore, our specific aim is to enhance the robustness of cooperative perception performance in the face of message loss stemming from unpredictable communication interruptions.

\begin{figure*}
\centering
\includegraphics[width=\linewidth]{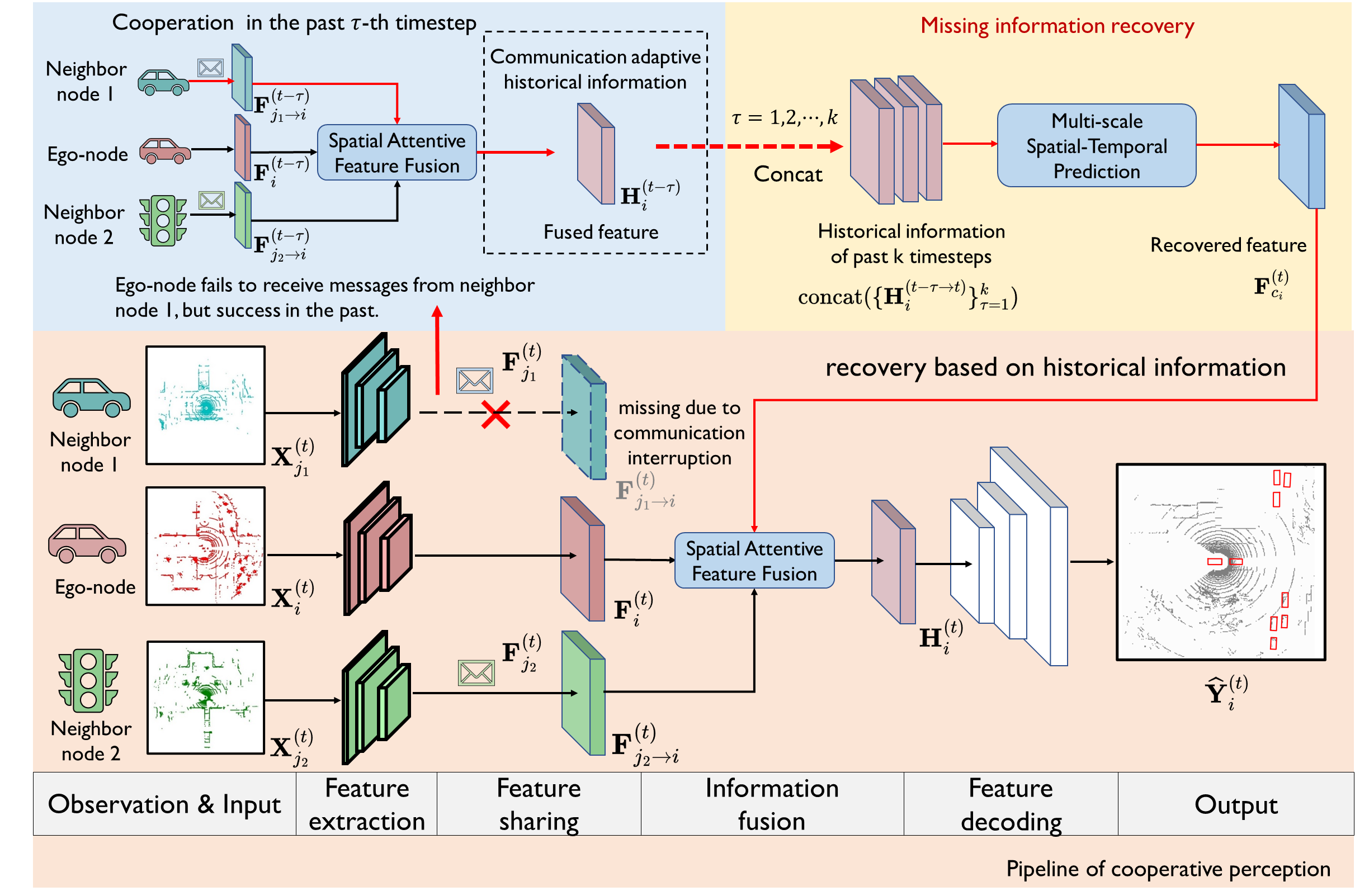}
\caption{\textbf{Overview of V2X-INCOP system.} The system adopts an intermediate cooperation paradigm and fuses features by spatial attentive feature fusion model. When communication interruption happens, V2X-INCOP recovers the missing information from the cooperation history through the missing information recovery process, which predicts the current state of the features of past timesteps from the communication-adaptive historical information with a multi-scale spatial-temporal prediction model. Then we regard the recovered feature as the feature from a pseudo cooperation node that can compensate for the missing information due to the communication interruption to complete the current feature fusion.
}
\label{fig:overview}%
\end{figure*} 

\section{Interruption-aware robust cooperative perception system}\label{sec:method}
In this section, we introduce our proposed V2X-INCOP system. First, we give an overview of the whole system. Second, we introduce the core component of the system: information recovery from historical information by a communication adaptive spatial-temporal attentive prediction model. Third, we introduce the training strategy, which further improves the accuracy and robustness of the system.

\subsection{Overview of the main system} \label{subsec:overview}

The proposed V2X-INCOP system leverages the intermediate cooperation paradigm, which exchanges intermediate features across different cooperation nodes. This is because intermediate features are much easier to compress compared with raw data and can be robust to noise compared with perception results, which can achieve a better trade-off between perception performance and communication cost. Although the proposed solution can be applied to any perception system regardless of the sensor types, we focus on LiDAR sensor-based 3D object detection for illustration of the proposed solution.When applying the system to other perception tasks, the only thing to do is replacing the corresponding feature extraction backbone for the sensor type and output heads for the task.

The whole processing pipeline is shown in Figure~\ref{fig:overview}. First, each cooperation node extracts the features of its raw sensor data and exchanges the features with neighbor nodes. When the ego node fails to receive some messages in the presence of communication interruption, it conducts missing information recovery from historical information to obtain recovered features. Then, ego features, received features and recovered features are aggregated by information fusion. Finally, the detection output is obtained by feature decoding. The details are as follows.

\textbf{Feature extraction.}
At each timestep $t$, node $a_i$ observes its surrounding environment to obtain local raw sensor data $\mathbf{X}_{i}^{(t)}$. We use a feature encoder $f_{\mathrm{encode}}$ to exact spatial features $\mathbf{F}_{i}^{(t)}$ of $\mathbf{X}_{i}^{(t)}$, that is,
\begin{equation*}
    \mathbf{F}_{i}^{(t)} = f_{\mathrm{encode}}(\mathbf{X}_{i}^{(t)}).
\end{equation*}  

For LiDAR-based 3D object detection, we use pointpillars~\cite{lang2019pointpillars} as an encoder, because it is computation-efficient and has advantages for realistic application.

\textbf{Feature sharing and transformation.}
Then, each node broadcasts its encoded features and poses in 3D space to neighboring nodes and receives this information from its neighbors. After receiving messages from other nodes, each node $a_i$ performs a transformation of features from other nodes into its own coordinate system, which is determined by their respective poses. This transformation is represented as:
\begin{equation*}
    \mathbf{F}_{j \to i}^{(t)} = \mathrm{\xi}_{j\to i}(\mathbf{F}_{j}^{(t)}).
\end{equation*}
Here, $\mathrm{\xi}_{j\to i}$ denotes the transformation principle reliant on the poses of both nodes, $a_i$ and $a_j$. After this coordinate transformation, all features are aligned within the same coordinate system.

\textbf{Missing information recovery.}
If communication interruption happens in the above process, node $a_i$ encounters a failure in receiving messages from certain node $a_j$ at timestep $t$. However, information relevant to the missing messages may have been received from node $a_j$ or other nodes at previous timesteps. We can thus recover the missing information due to interruptions from historical information by estimating the current state of the fused features at each previous timestep. For example, node $a_i$ fails to receive $\mathbf{F}_{j}^{(t)}$, possibly causing a misdetection of an object $o_i$ observed by node $a_j$. Forturnately, the ego node $a_i$ has received the features $\mathbf{F}_{j}^{(t-\tau)}$ containing information of the object $o_i$ from node $a_j$ at timestep $t-\tau$, and this information has been integrated into the feature $\mathbf{H}_{i}^{(t-\tau)}$ during the feature fusion at timestep $t-\tau$. We call the fused features at previous timesteps in cooperation history \textit{historical information}. Therefore, we can estimate the current state of object $o_i$ within the features from historical information and thus recover the missing information due to communication interruptions. Inspired by this, we propose a multi-scale spatial-temporal prediction model $f_{\mathrm{predict}}$, which leverages the historical information at past $k$ timesteps $\mathrm{concat}(\{\mathbf{H}_{i}^{(t-\tau)}\}_{\tau = 1}^{k})$ to recover the current missing information $\mathbf{F}_{c_i}^{(t)}$. That is,
\begin{equation*}
    \mathbf{F}_{c_i}^{(t)} = f_{\mathrm{predict}}(\mathrm{concat}(\{\mathbf{H}_{i}^{(t-\tau)}\}_{\tau = 1}^{k})).
\end{equation*}

The details of the multi-scale spatial-temporal prediction model will be introduced in the next subsection.

\begin{figure}
\centering
\includegraphics[width=\linewidth]{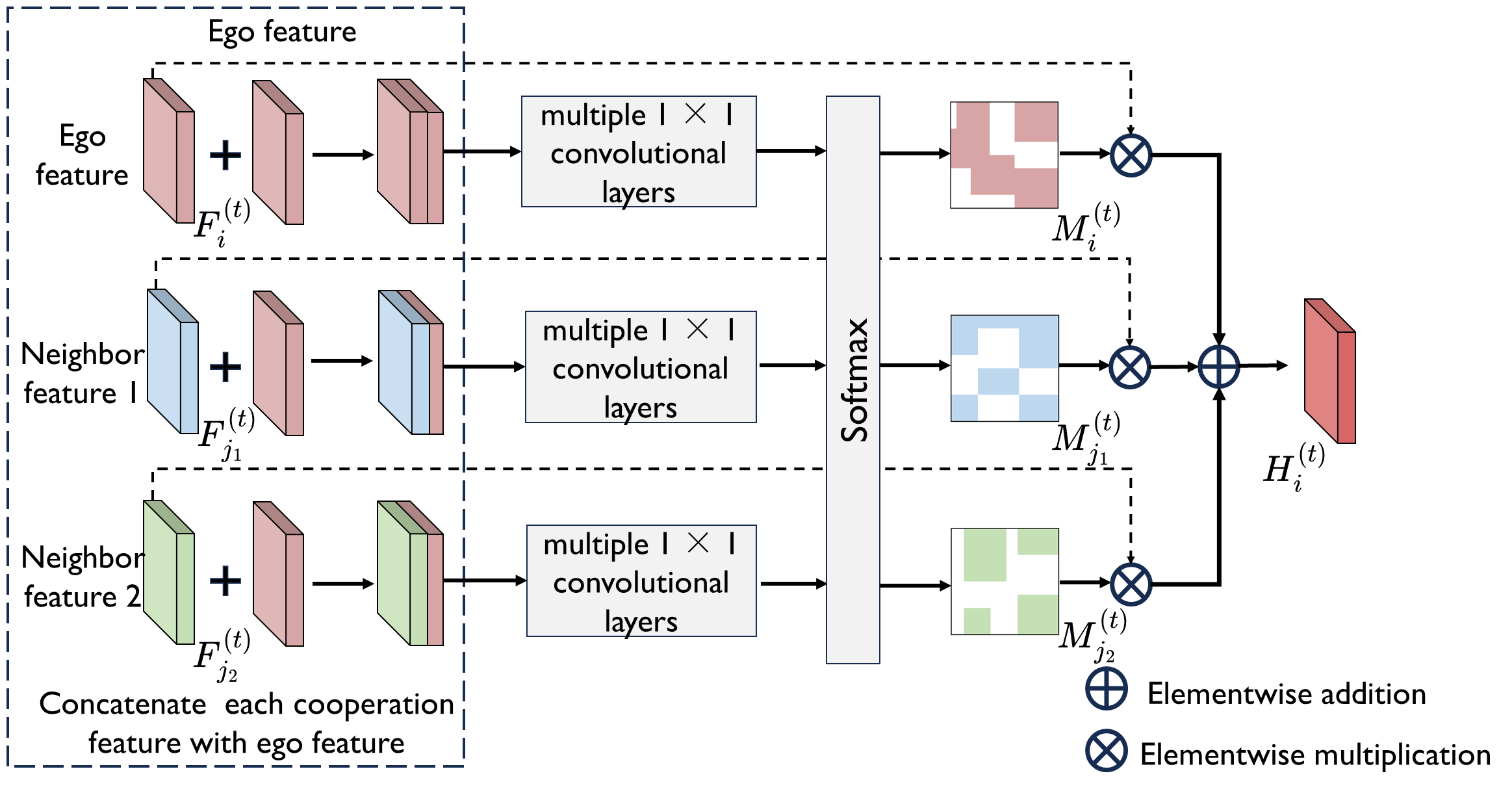}
\caption{\textbf{Spatial attentive feature fusion (SAFF) model.} The SAFF model first computes the spatial attention weight of each feature and then fuse them according to the weight, which captures the informative regions for the ego node of each cooperation feature and obtains the fused feature adaptive to the communication condition. }
\label{fig:mask}%
\end{figure} 

\textbf{Information fusion.}
Following missing information recovery, we consider the recovered feature $\mathbf{F}_{c_i}^{(t)}$ as originating from a pseudo cooperation node, capable of compensating for the information gaps caused by communication interruption. We employ a spatial attentive feature fusion (SAFF) model denoted as $f_{\mathrm{fuse}}$ to fuse the features from all nodes, which includes the pseudo node, resulting in a fused feature $\mathbf{H}_{i}^{(t)}$. That is,
\begin{equation*}
    \mathbf{H}_{i}^{(t)} = f_{\mathrm{fuse}}(\{\mathbf{F}_{j \to i}^{(t)}\}_{j \in \{i, c_i\} \cup \mathcal{R}_{i}^{(t)}}). \tag{*}
    \label{eq:fusion}
\end{equation*}

The details of the SAFF model will be introduced in the next subsection.

\textbf{Feature decoding.}
Finally, we employ a decoder denoted as $f_{\mathrm{decode}}$ to decode the fused features, resulting in the perception results $\widehat{\mathbf{Y}}_{i}^{(t)}$ for each node. That is,
\begin{equation*}
    \widehat{\mathbf{Y}}_{i}^{(t)} = f_{\mathrm{decode}}(\mathbf{H}_{i}^{(t)}).
\end{equation*}

In terms of 3D object detection, each result consists of a pair of classification and regression results, obtained by a classification head and a regression head, respectively. The classification result is predicted probabilities for foreground-background categories and the regression results are regressed parameters of bounding boxes, including 3D coordinates of box center, length, width, height, and yaw angle of the box.

\subsection{Information recovery from historical information}
This subsection introduces the details of missing information recovery and information fusion. We propose a communication adaptive multi-scale spatial-temporal prediction model to achieve missing information recovery from historical information and propose a SAFF model to fuse features from different cooperation nodes. We will introduce its details in three parts: communication adaptive historical information, multi-scale spatial-temporal prediction model, and knowledge distillation from oracle.

\subsubsection{Communication adaptive historical information}
% \textbf{Communication adaptive historical information.}
The input of the prediction model is \textit{communication adaptive historical information}, which is obtained by fusing messages from different nodes at each past timestep with SAFF model, shown in Figure~\ref{fig:mask}. In this part, we first introduce the proposed SAFF model, whose functionality is to attentively fuse features with different communication interruption conditions, and then we get the \textit{communication adaptive historical information}. Note that the SAFF model is not only used to generate historical information but also used to achieve feature fusion at all timesteps.

The functionality of SAFF model is to fuse features according to the spatial attention of each feature.
As different cooperation nodes have distinct locations and views, the features from these nodes inherently take information from diverse spatial regions. Consequently, different spatial cells within these features hold distinct levels of significance to the ego node. In pursuit of acquiring the most dependable information, it becomes imperative to assign diverse weights to different spatial cells. The SAFF model serves to amplify the informative cells while simultaneously attenuating the irrelevant and noisy cells during the fusion process. The details of SAFF model are introduced in the following.

Given the ego feature $\mathbf{F}_{i}^{(t)}$ and neighbor features $\mathbf{F}_{j \to i}^{(t)}$, SAFF model first computes the spatial attention weight of each feature and then fuse them according to the weight.
 To compute the weight, we concatenate the ego feature $\mathbf{F}_{i}^{(t)}$ with each feature from neighbor node $a_j$, $\mathbf{F}_{j \to i}^{(t)}$, along the channel dimension and use multiple 1 × 1 convolutional layers to progressively reduce the number of channels to 1. The above calculation process is denoted by $f_{\mathrm{mask}}$. Following this, we apply a softmax function to every pixel over all features, resulting in the generation of a spatial mask, i.e. the spatial attention weight $\mathbf{M}_{j}^{(t)}$ of each feature $\mathbf{F}_{j}^{(t)}$. That is,
\begin{equation*}
    \mathbf{M}_{j}^{(t)} = \mathrm{softmax}(f_{\mathrm{mask}}(\mathrm{concat}(\mathbf{F}_{i}^{(t)}, \mathbf{F}_{j \to i}^{(t)}))),
\end{equation*}
for all $ j \in \{i, c_i\} \cup \mathcal{R}_{i}^{(t)}$. The multiple 1 × 1 convolutional layers are used to capture the most informative region for the ego node from the concatenation of the cooperation feature and the ego feature, and the softmax function is used to output normalized weights for feature fusion.

With the computed spatial attention weights, we perform feature fusion by employing weighted averaging and obtain the fused feature of node $a_i$, that is,
\begin{align*}
    \mathbf{H}_{i}^{(t)} = &f_{\mathrm{fuse}}(\{\mathbf{F}_{j \to i}^{(t)}\}_{j \in \{i\} \cup \mathcal{R}_{i}^{(t)}}) \\
    = &\sum_{j \in \{i, c_i\} \cup \mathcal{R}_{i}^{(t)}}
    {\mathbf{M}_{j}^{(t)} \odot \mathbf{F}_{j \to i}^{(t)}},
\end{align*}
where $f_{\mathrm{fuse}}$ denotes the SAFF model and $\odot$ is an element-wise multiplication.

In this way, the output of the SAFF model is adaptive to the communication state and captures the most important information to different nodes at each timestep. We concatenate the fused features at the past $k$ timesteps to obtain the \textit{communication adaptive historical information} for node $a_i$, $\mathrm{concat}(\{\mathbf{H}_{i}^{(t-\tau)}\}_{\tau = 1}^{k}$. Therefore, the communication adaptive historical information carries the information for the following recovery.

\begin{figure*}[t]
	\centering
	\subfigure[Architecture of multi-scale spatial-temporal prediction model.]{
		\begin{minipage}[b]{0.45\textwidth}
			\includegraphics[width=1\textwidth]{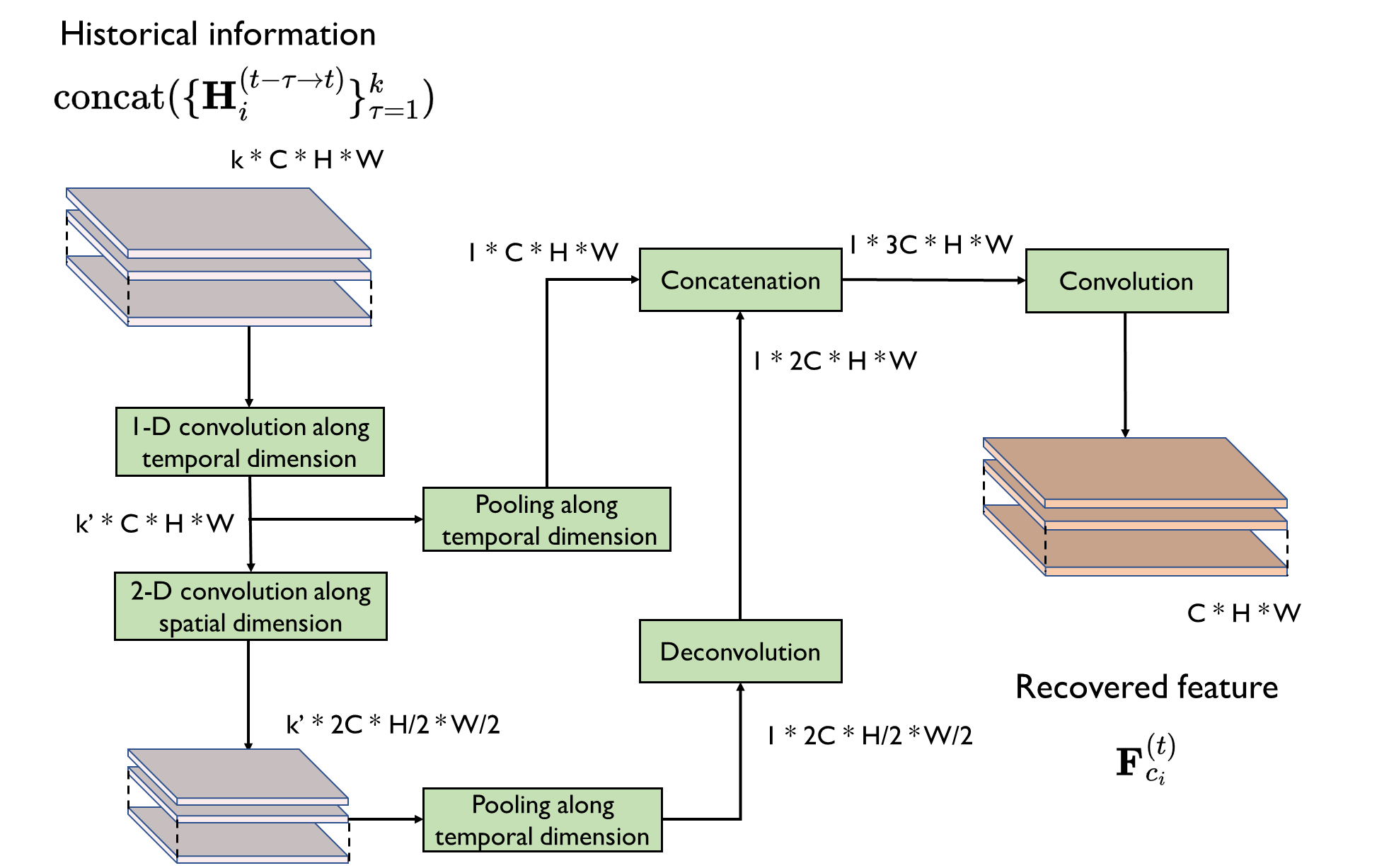}
		\end{minipage}
		% \label{fig:pre_arc}
	}
        \subfigure[Illustration of knowledge distillation framework.]{
    		\begin{minipage}[b]{0.48\textwidth}
   		 	\includegraphics[width=1\textwidth]{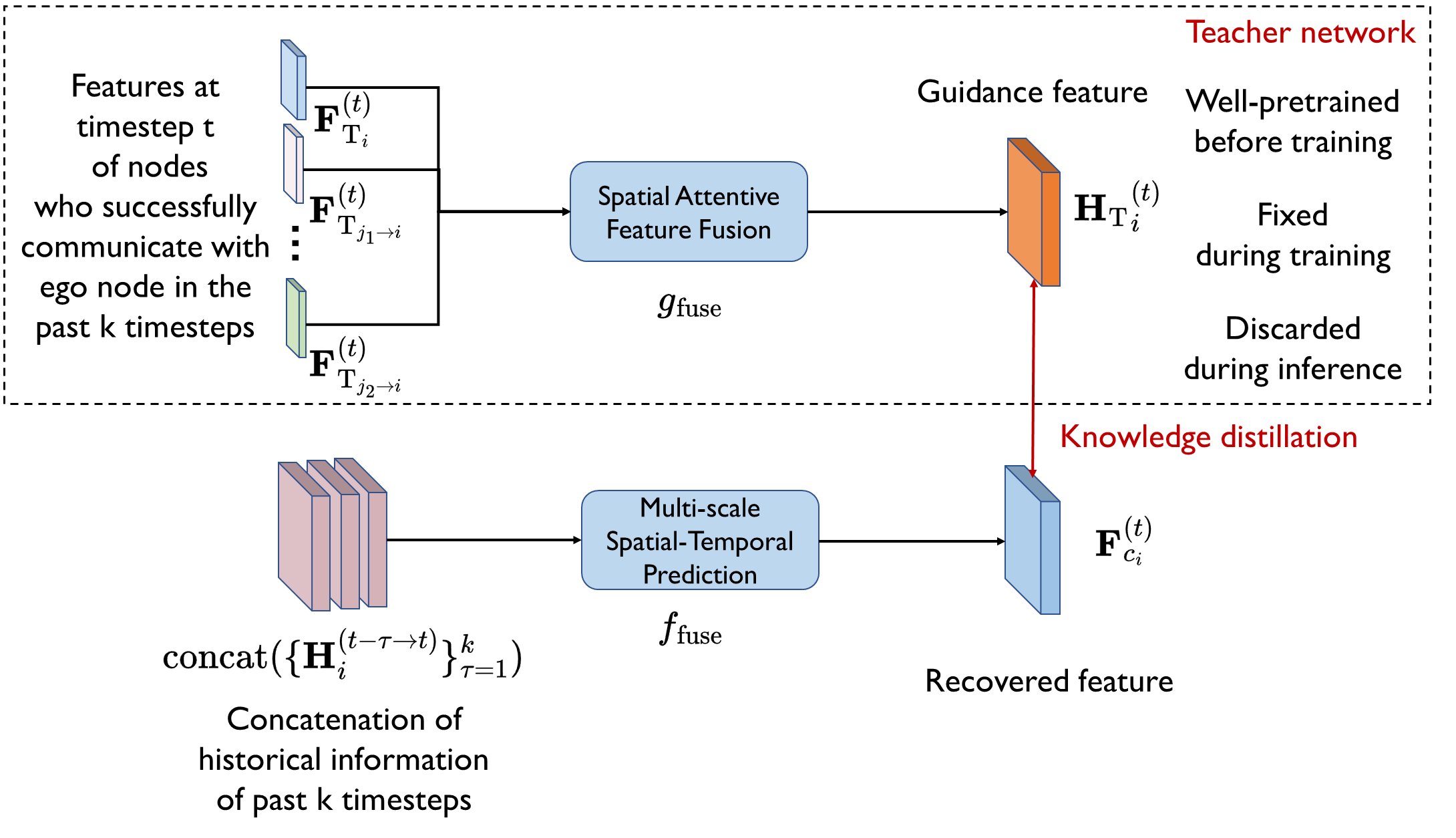}
    		\end{minipage}
		% \label{fig:kd_framework}
    	}
    \caption{\textbf{Multi-scale spatial-temporal prediction model.} (a) The model first extracts multi-scale spatial-temporal features of historical information and then predicts the recovered feature. (b) The model is given explicit supervision during training with knowledge distillation from a teacher model without communication interruptions.}
    \label{fig:prediction model}
\end{figure*}

\subsubsection{Multi-scale spatial-temporal prediction model}\label{subsec:msspm}

The functionality of the multi-scale spatial-temporal prediction model is to estimate the current state of the features at past timesteps given the \textit{communication adaptive historical information} to recover the missing information due to communication interruptions. 
As mentioned above in the overview, some information relevant to the missing
messages have been integrated into \textit{communication adaptive historical information} in the cooperation history, we can recover the missing information due to communication interruptions by estimating the current state of the features at past timesteps.
Moreover, the \textit{communication adaptive historical information} carries spatial-temporal information from multiple cooperation nodes and multiple timesteps where objects have multiple sizes, the prediction model needs to extract the multi-scale spatial-temporal features for comprehensive prediction. The following part will introduce the design of the multi-scale spatial-temporal prediction model, whose architecture is shown in Figure~\ref{fig:prediction model}(a).

Given the \textit{communication adaptive historical information} for node $a_i$, $\mathrm{concat}(\{\mathbf{H}_{i}^{(t-\tau)}\}_{\tau = 1}^{k}$, We first extract multi-scale spatial-temporal features of historical information with a spatial-temporal pyramid network. This network extracts features through two groups of networks. Each group of networks is mainly composed of two 2D convolution layers in the spatial dimension and a 1-D convolution layer in the time dimension. For spatial dimension, 2D convolution layers are applied to downsample the input feature to extract the spatial features. For the temporal dimension, the 1-D convolution layer is applied to gradually reduce the temporal resolution. After obtaining multiple levels of spatial-temporal features, we fuse them with skip connections, which concatenate the lower level of features with the upsampled upper level of features. 
In this way, we obtain the features with different spatial resolutions and dynamic information in the temporal dimension, which takes enough information for accurate and complete recovery.

Subsequently, two convolution layers are applied to generate the output, denoted as $\mathbf{F}_{c_i}^{(t)}$, with the size matching those of the feature to be fused, as defined in Eq.~\ref{eq:fusion}, where $c_i$ represents the index of the predicted feature. The whole prediction process can be formulated as follows,
\begin{equation*}
    \mathbf{F}_{c_i}^{(t)} = f_{\mathrm{predict}}(\mathrm{concat}(\{\mathbf{H}_{i}^{(t-\tau \to t)}\}_{\tau = 1}^{k})),
\end{equation*}
where $f_{\mathrm{predict}}$ denotes the multi-scale spatial-temporal prediction model and $\{\mathbf{H}_{i}^{(t-\tau \to t)}\}_{\tau = 1}^{k})$ denotes the \textit{communication adaptive historical information} for node $a_i$. Therefore, the model predicts the features representing the current state of historical information based on their spatial-temporal features.

\subsubsection{Knowledge distillation from oracle}
% \textbf{Knowledge distillation from oracle.} 

The functionality of knowledge distillation in the proposed method is to provide the ground truth of the fused features without communication interruptions at the current timestep to the prediction model during the training stage, guiding the model to estimate the current state of the historical information and improving its performance. When we solely supervise the model with a detection loss for object detection results,  there is a deficiency in explicit and direct supervision for the output of the prediction model, leading to a lack of an explicit optimization target. This limitation hinders overall performance. Fortunately, we can obtain the fused feature without communication interruption at the current timestep during the training stage, which serves as the ground truth of the prediction model (\textit{the oracle}). We propose to leverage knowledge distillation to provide this supervision with a teacher model fusing features without communication interruption, which can guide the prediction model to recover the missing information from historical information. This part will introduce how we train the multi-scale spatial-temporal prediction model with knowledge distillation.

Knowledge distillation (KD)~\cite{hinton2015distilling} is a teacher-student framework that leverages a pre-trained teacher network to guide the training of a student network by minimizing the difference between the outputs of the teacher and the student network.

The knowledge distillation framework used in V2X-INCOP is shown in Figure~\ref{fig:prediction model}(b). In the system, the student network is the prediction model mentioned in subsection~\ref{subsec:msspm}. The teacher network is a virtual cooperative perception network \textit{without} communication interruption, which fuses the features at the current timestep from the cooperative nodes who are involved in the cooperation history.
The fused feature is a guidance feature that guides the training of the prediction model, leading the model to estimate the current state of the past features from historical information.
The teacher network is composed of an encoder $g_{\mathrm{encoder}}$, an information fusion model $g_{\mathrm{fuse}}$, and a decoder model $g_{\mathrm{decoder}}$, and the architecture of these models are the same as that of the models in the cooperative perception system mentioned in subsection \ref{subsec:overview}, that is, $f_{\mathrm{encoder}}$, $f_{\mathrm{fuse}}$ and $f_{\mathrm{decoder}}$. 

The teacher network is first well-trained with the assumption of ideal communication.
Then we can obtain a guidance feature by feeding the teacher network with point clouds of nodes in $\mathcal{P}_{i}^{t}$ at the current timestep, where $\mathcal{P}_{i}^{t}$ is the set of nodes from who node $a_i$ has successfully received messages at the past $k$ timesteps before timestep $t$. We compute the guidance feature of the prediction model as follows:
\begin{equation*}
    {\mathbf{F}_{\mathrm{T}}}_{i}^{(t)} = g_{\mathrm{encode}}(\mathbf{X}_{i}^{(t)}),
\end{equation*}  
\begin{equation*}
    {\mathbf{F}_\mathrm{T}}_{j \to i}^{(t)} = \mathrm{\xi}_{j\to i}({\mathbf{F}_\mathrm{T}}_{j}^{(t)}),
\end{equation*}
\begin{equation*}
    {\mathbf{H}_\mathrm{T}}_{i}^{(t)} = g_{\mathrm{fuse}}(\{{\mathbf{F}_{\mathrm{T}}}_{i}^{(t)}\}_{j \in \{i\} \cup \mathcal{P}_{i}^{(t)}}),
\end{equation*}
where subscript $\mathrm{T}$ refers to the variable produced by the teacher network.

The guidance feature can be regarded as the ground truth of the output of the prediction model, because if the prediction model outputs the same feature as the guidance feature, it has recovered all information needed to be recovered based on the feature received in the past.

In the training stage, we push the output of the prediction model to approach the guidance feature with a KD loss.
In the inference stage, the teacher network is discarded because messages from some nodes are inaccessible due to interruption. However, the prediction model has learned to extract more important spatial-temporal features and output better prediction with the guidance of the teacher network.

\subsection{Training strategy}

\subsubsection{Curriculum learning}

In order to enable the model to address varying communication conditions, we introduce a mechanism where the probability of communication interruption at each iteration is stochastically sampled from the range of $(0,1)$ during the training phase, which may result in unstable training and affect the performance, because the training loss exhibits a strong correlation with the probability of interruption. For example, the training loss increases with a larger interruption probability because more missing cooperation messages cause more perception errors.

To address this challenge, we employ the concept of curriculum learning~\cite{bengio2009curriculum} wherein the learning process initially commences with easy examples and progressively escalates the task difficulty. It is thought that cooperative perception with low interruption probabilities is a simpler case for the prediction and detection models in our system. Hence, we initiate the training process with a low probability of interruption and systematically expand the interruption probability range. When the interruption probability is low at the beginning of the training, the fused features can contain more complete and useful information so that the detection model and the prediction model can obtain better features as input to generate results closer to the ground truth, making the training process have a good initialization. After the training in the early stage with simple cases, the system acquires basic detection and prediction capabilities. The range of communication interruption probabilities is then gradually increased, and the system learns to predict and detect with some missing information. 
 Therefore, the training of the system can have an easier start and gradually focus on harder samples. Finally, the system can adapt to various communication conditions. In this way, the training process can achieve better convergence and the well-trained model has a better generalization performance under different communication interruption conditions.

\subsubsection{Training loss}

The proposed V2X-INCOP system is optimized by minimizing the loss function associated with the perception task itself and a knowledge distillation loss, which facilitates its seamless application and transferability across a spectrum of perception tasks. In this work, we focus on the task of Lidar-based cooperative 3D object detection. So the loss function is composed of a detection loss and a KD loss.

\textbf{Detection loss.} The detection loss is composed of a classification loss and a regression loss. The results of detection $\widehat{\mathbf{Y}}$ consists of a classification result $\widehat{\mathbf{Y}}_{\rm cls}$ and a location result $\widehat{\mathbf{Y}}_{\rm loc}$. Let $\mathbf{Y}$ represent the ground truth for object detection which comprises classification ground truth denoted as $\mathbf{Y}_{\rm cls}$ and location ground truth denoted as $\mathbf{Y}_{\rm loc}$. 
The classification loss is a binary cross-entropy loss between $\widehat{\mathbf{Y}}_{\rm cls}$ and $\mathbf{Y}_{\rm cls}$. The regression loss is a smooth L1 loss between  $\widehat{\mathbf{Y}}_{\rm loc}$ and $\mathbf{Y}_{\rm loc}$. 
The detection loss function can be formulated as:
\begin{align*}
\mathcal{L}_{\rm Det}(\widehat{\mathbf{Y}},\mathbf{Y})=~~~~~~~~~~~~~~~~~~~~~~~~~~~~~~~~~~~~~~~~~~~~~~~~~\\
\alpha\mathrm{L}_{\rm CE}\left(\widehat{\mathbf{Y}}_{\rm cls},\mathbf{Y}_{\rm cls}\right)+ \beta\mathrm{L}_{\rm smooth-L1}\left(\widehat{\mathbf{Y}}_{\rm loc},\mathbf{Y}_{\rm loc}\right),
\end{align*}
where $\alpha$ and $\beta$ are coefficients for balancing weights.

\textbf{KD loss.} When training the multi-scale spatial-temporal prediction model, we leverage a KD loss $\mathcal{L}_{\rm KD}$ to push the output of prediction model $\mathbf{F}_{c_i}^{(t)}$ to approach the guidance feature ${\mathbf{H}_\mathrm{T}}_{i}^{(t)}$. The loss is the Kullback-Leibler (KL) divergence of the distribution of the output of the prediction model from the guidance feature, that is,
\begin{align*}
\mathcal{L}_{\rm KD}(  \mathbf{F}_{c_i}^{(t)}, {\mathbf{H}_\mathrm{T}}_{i}^{(t)} ) = ~~~~~~~~~~~~~~~~~~~~~~~~~~~~~~~~~~~~~~~~ \\
\sum^{w \times h}_{n=1}  D_{\rm KL} \left(\sigma \left(  \left({ \mathbf{F}_{c_i}^{(t)}} \right)_n  \right)|| \sigma \left( \left({\mathbf{H}_\mathrm{T}}_{i}^{(t)} \right)_n \right)  \right),
\end{align*}
where $D_{KL}(\cdot||\cdot)$ denotes the Kullback-Leibler (KL) divergence of two distributions and $\sigma(\cdot)$ indicates the softmax operation along the channel dimension of the feature. $w$ and $h$ denote the width and height of the feature, respectively. $\left({\mathbf{F}_{c_i}^{(t)}} \right)_n$ and $\left({\mathbf{H}_\mathrm{T}}_{i}^{(t)} \right)_n$ denote the feature vectors at the $n$th cell of ${  \mathbf{F}_{c_i}^{(t)}}$ and ${\mathbf{H}_\mathrm{T}}_{i}^{(t)}$, respectively. 

Finally, the total loss of V2X-INCOP system is then obtained as $\mathcal{L}_{\rm Total} = \mathcal{L}_{\rm Det} + \gamma\mathcal{L}_{\rm KD}$, where $\gamma$ is a coefficient for balancing weights.

\begin{figure*}[t]
	\centering
	\subfigure[V2X-Sim]{
		\begin{minipage}[b]{0.3\textwidth}
			\includegraphics[width=1\textwidth]{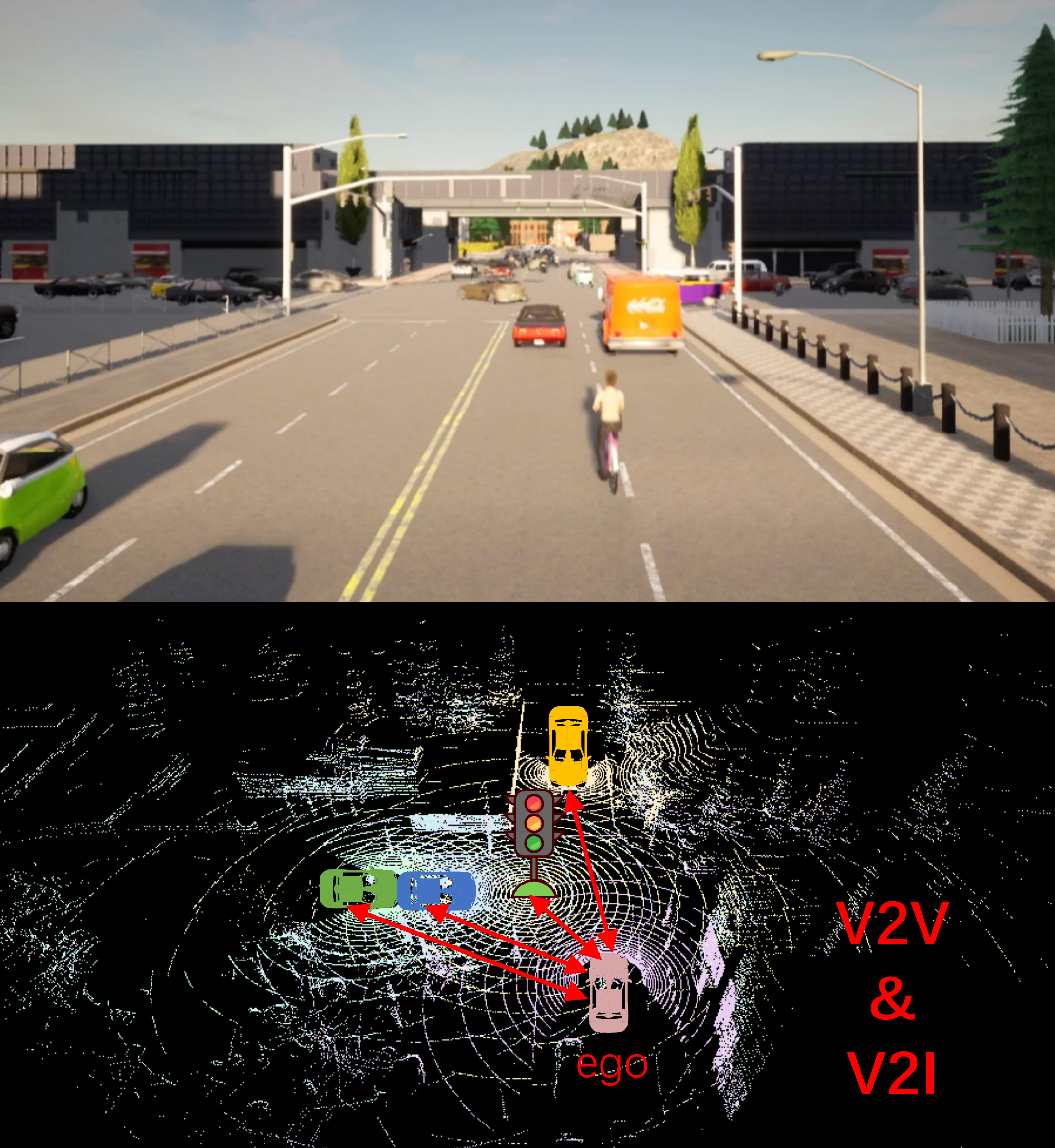}
		\end{minipage}
		% \label{fig:issue}
	}
    	\subfigure[OPV2V]{
    		\begin{minipage}[b]{0.3\textwidth}
   		 	\includegraphics[width=1\textwidth]{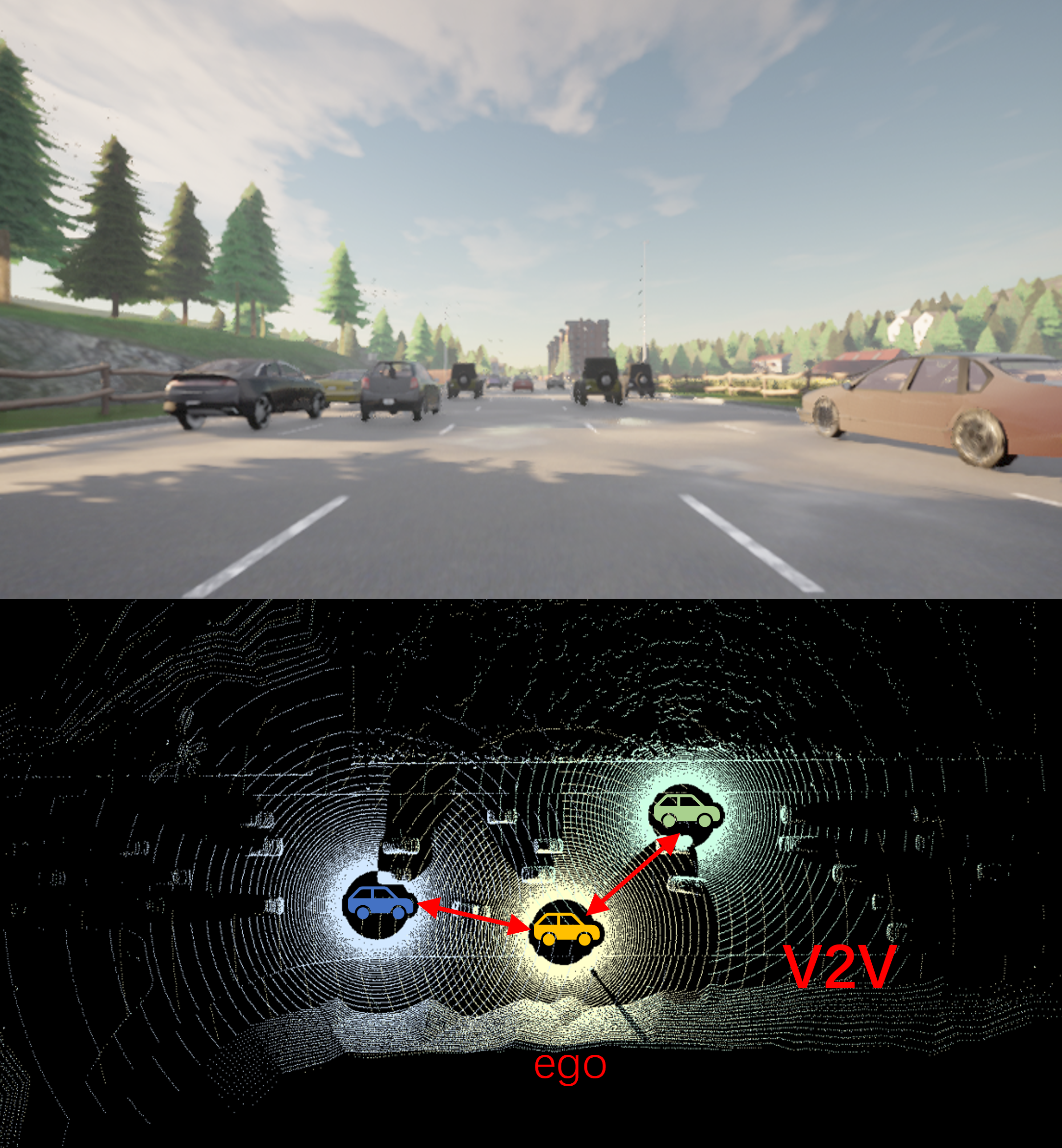}
    		\end{minipage}
		% \label{fig:performance degrade}
    	}
        \subfigure[DAIR-V2X]{
    		\begin{minipage}[b]{0.3\textwidth}
   		 	\includegraphics[width=1\textwidth]{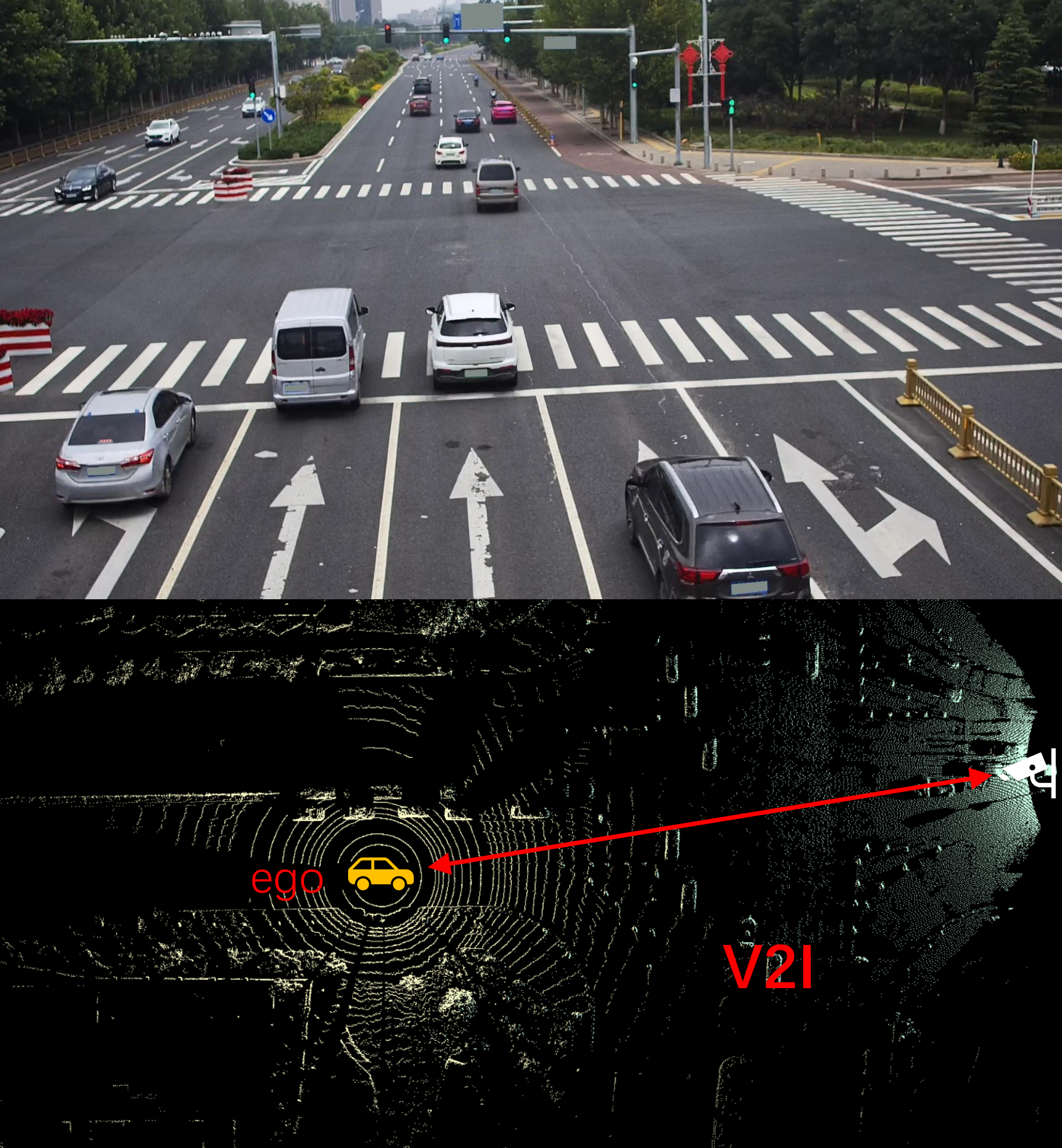}
    		\end{minipage}
		% \label{fig:performance degrade}
    	}
    \caption{\textbf{Visualization of datasets.} The two rows show the exemplar scenarios from datasets V2X-Sim, OPV2V and Dair-V2X in form of images and point clouds, respectively. V2X-Sim includes V2V and V2I scenarios in crowded downtown and suburb highway. OPV2V includes V2V scenarios with diverse city traffic configurations. Dair-V2X collects data from real-world vehicle-infrastructure cooperation scenarios. }
    \label{fig:dataset}
\end{figure*}

\section{Evaluation}\label{sec:exp}

In this section, we introduce the details of experiments, including datasets, evaluation metrics, implementation, quantitative and qualitative evaluation, ablation study and discussion.

\subsection{Dataset}

In this work, we validate our proposed methods on the task of LiDAR-based 3D object detection. We conduct our experiments on three public cooperative perception datasets, covering both simulation and real-world scenarios, as described in the following. Figure~\ref{fig:dataset} shows the exemplar scenarios from the three datasets in the form of images and LiDAR point clouds.

\textbf{V2X-Sim dataset.}
V2X-Sim~\cite{li2022v2x} is a cooperative perception dataset simulated by CARLA~\cite{dosovitskiy2017carla} and SUMO~\cite{krajzewicz2012recent}. There are synchronized sensor streams from both vehicles and roadside units, which enables V2X scenarios. It includes 10K frames of 3D point clouds and 501K annotated 3D boxes. Each frame contains 2$\sim$5 vehicles and one roadside unit as cooperation nodes. 8K/1K/1K frames are selected as training/validation/testing samples, respectively. The perception range is $x \in [-32m, 32m], y \in [-32m, 32m]$.

\textbf{OPV2V dataset.}
OPV2V~\cite{xu2022opencood} is a V2V perception dataset simulated by CARLA and OpenCDA~\cite{10045043}. It includes 11,464 frames of 3D point clouds and 232,913 annotated 3D boxes. Each frame contains 2$\sim$7 vehicles as cooperation nodes. 6,374/2,920/2,170 frames are selected as training/validation/testing samples, respectively. The perception range is $x \in [-140m, 140m], y \in [-40m, 40m]$.

\textbf{DAIR-V2X dataset.}
DAIR-V2X~\cite{yu2022dair} is the first real-world vehicle-infrastructure cooperative perception dataset. All frames are captured from real scenarios with 3D annotations. We use the cooperation part of this dataset, DAIR-V2X-C. We select 3,633/1,408 frames as training/testing samples, respectively. The perception range is $x \in [-100.8m, 100.8m], y \in [-40m, 40m]$.

\subsection{Metrics and implementation}

To evaluate the performance of cooperative perception with communication interruption, we test the detection results in Birds-Eye-View with different V2X communication PDRs $(0\sim90\%)$.
The detection results are evaluated by Average Precision (AP) at Intersection-over-Union (IoU) thresholds of 0.50 and 0.70.

\begin{figure*}[t]
	\centering
	\subfigure[\parbox{0.3\textwidth}{\centering Performance in AP@IOU=0.5 \\ in V2V scenario}]{
		\begin{minipage}[b]{0.30\textwidth}
			\includegraphics[width=1\textwidth]{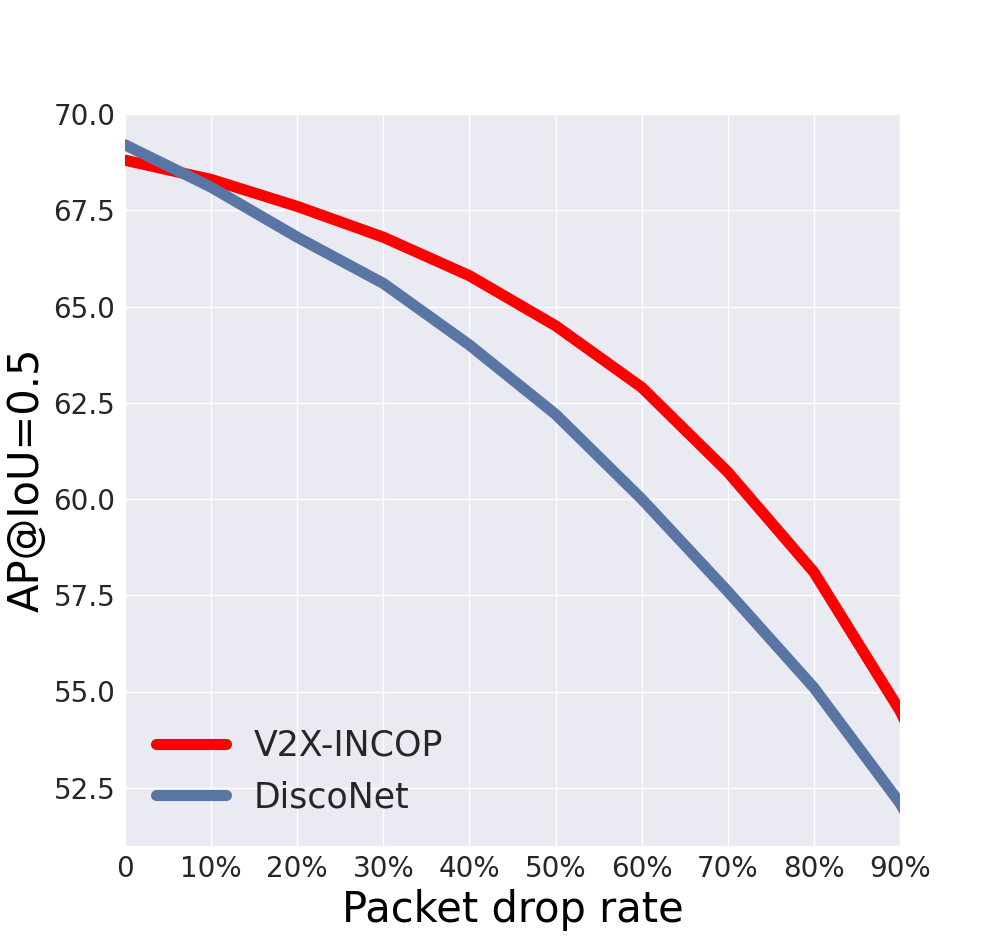}
		\end{minipage}
		% \label{fig:issue}
            % \caption{Performance in AP@IOU 0.5 in V2V scenario}
	}
    	\subfigure[\parbox{0.3\textwidth}{\centering Performance in AP@IOU=0.5 \\ in V2I scenario}]{
    		\begin{minipage}[b]{0.30\textwidth}
   		 	\includegraphics[width=1\textwidth]{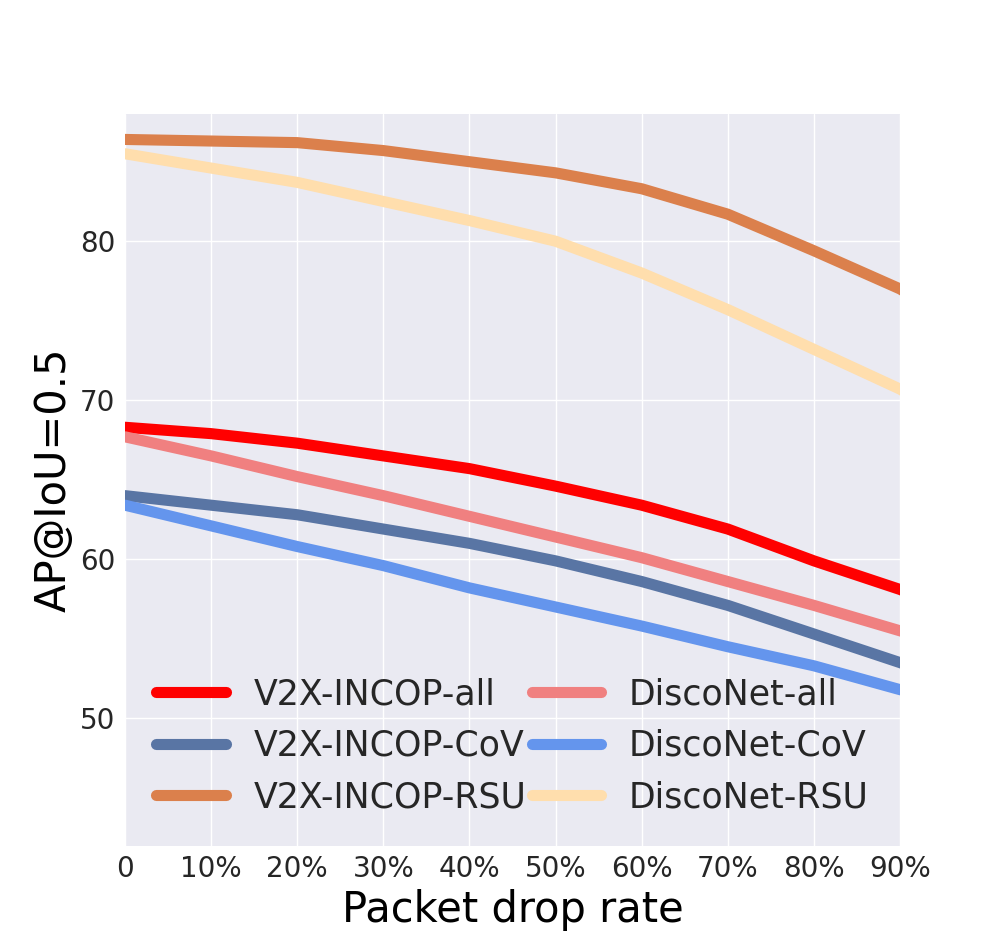}
    		\end{minipage}
		% \label{fig:performance degrade}
    	}
            \subfigure[\parbox{0.3\textwidth}{\centering Performance in AP@IOU=0.5 \\ in V2X scenario}]{
    		\begin{minipage}[b]{0.30\textwidth}
   		 	\includegraphics[width=1\textwidth]{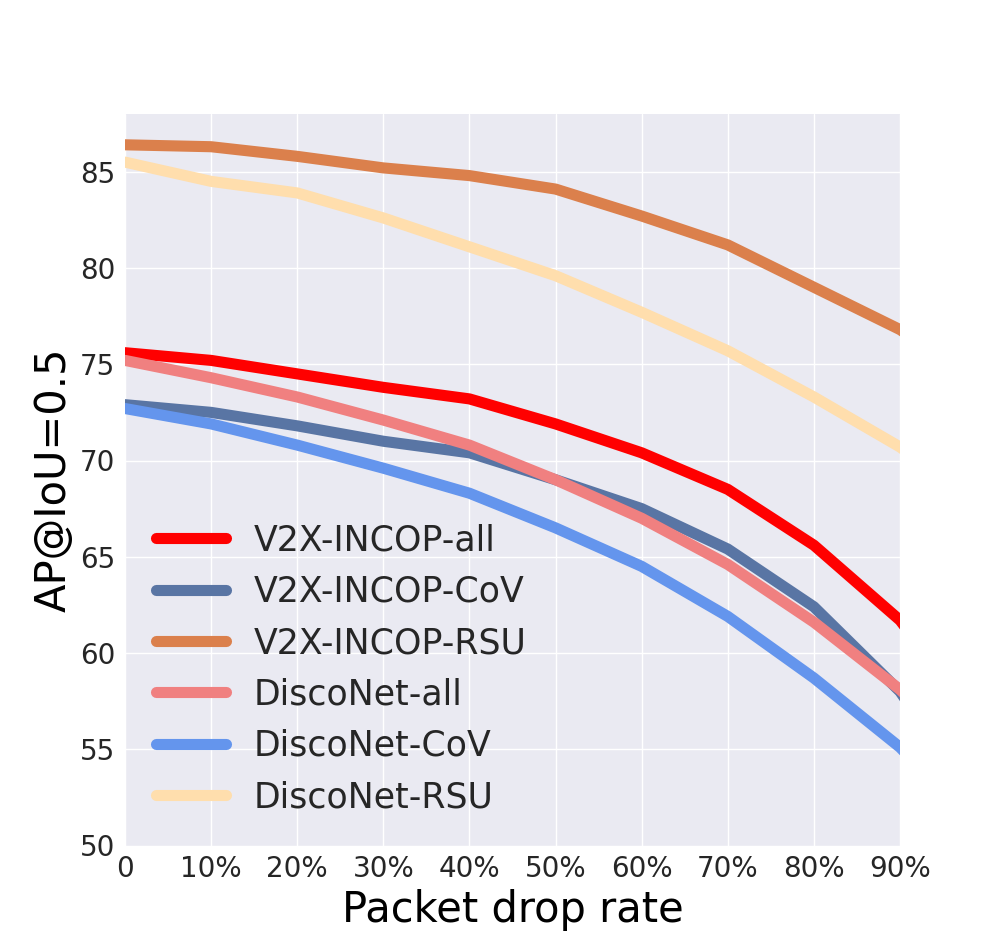}
    		\end{minipage}
		% \label{fig:performance degrade}
    	}

     	\subfigure[\parbox{0.3\textwidth}{\centering Performance in AP@IOU=0.7 \\ in V2V scenario}]{
		\begin{minipage}[b]{0.30\textwidth}
			\includegraphics[width=1\textwidth]{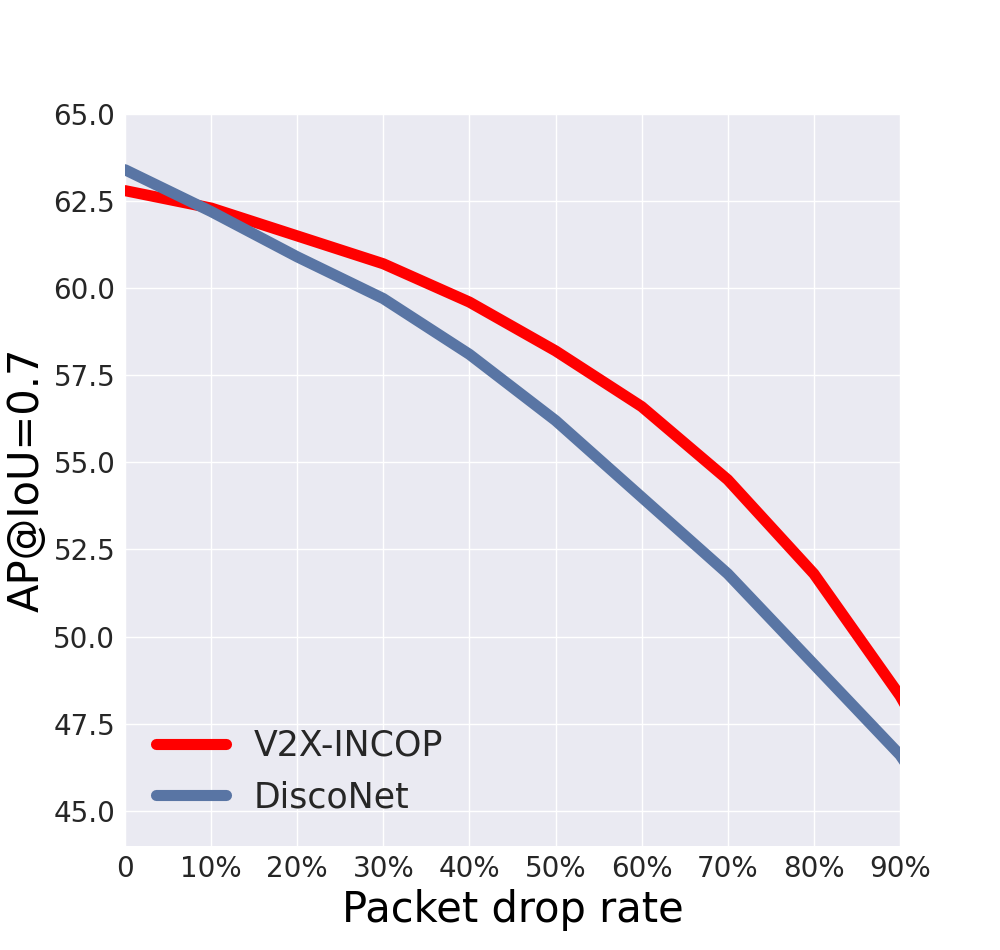}
		\end{minipage}
		% \label{fig:issue}
	}
    	\subfigure[\parbox{0.3\textwidth}{\centering Performance in AP@IOU=0.7 \\ in V2I scenario}]{
    		\begin{minipage}[b]{0.30\textwidth}
   		 	\includegraphics[width=1\textwidth]{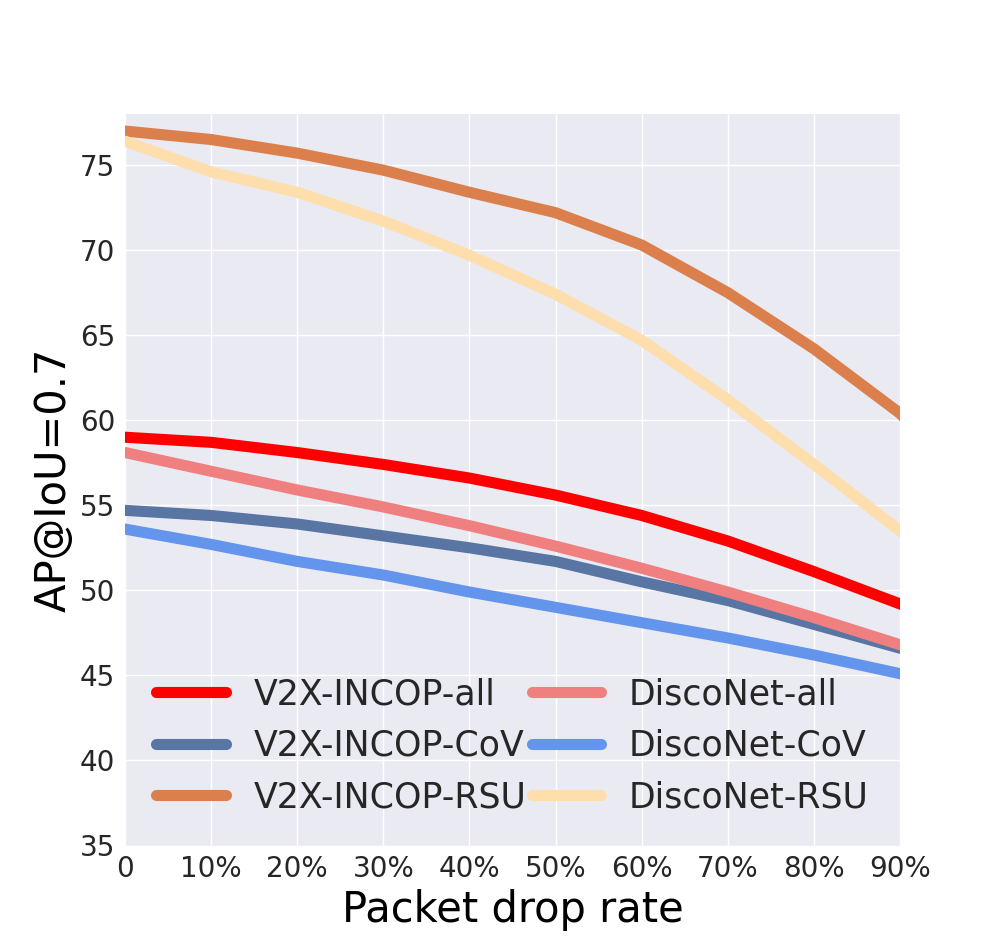}
    		\end{minipage}
		% \label{fig:performance degrade}
    	}
            \subfigure[\parbox{0.3\textwidth}{\centering Performance in AP@IOU=0.7 \\ in V2X scenario}]{
    		\begin{minipage}[b]{0.30\textwidth}
   		 	\includegraphics[width=1\textwidth]{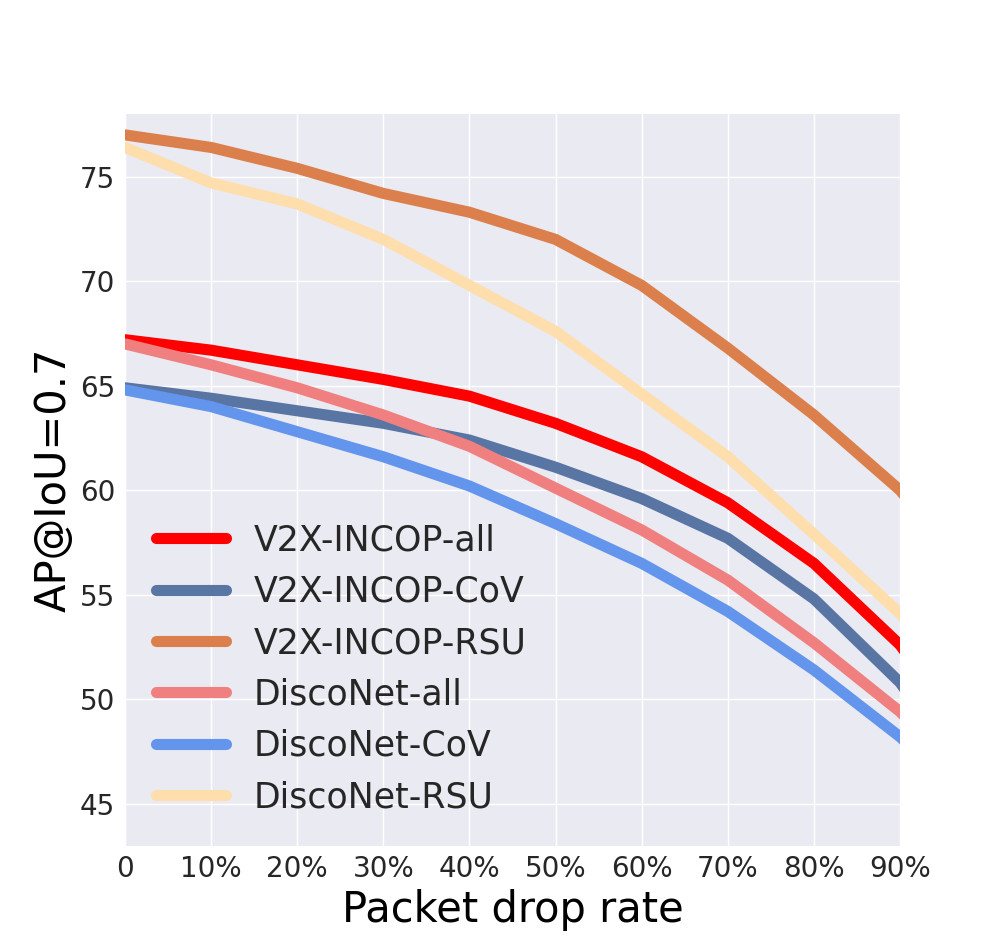}
    		\end{minipage}
		% \label{fig:performance degrade}
    	}
    \caption{\textbf{Detection performance with different packet drop rates on V2X-Sim dataset.} The performance is evaluated in three scenarios: V2V, V2I, and V2X, and measured by the average precision (AP) at intersection over union (IOU) thresholds of 0.5 and 0.7. The horizontal axis represents packet drop rates and the vertical axis represents the average precision of detection results. The results show that the proposed V2X-INCOP outperforms DiscoNet at most packet drop rates in all three scenarios. The robustness to communication interruption of both vehicle nodes and roadside nodes can be improved by V2X-INCOP. }
    \label{fig:v2x-sim}
\end{figure*}

\begin{figure*}[t]
	\centering
	\subfigure[Performance on OPV2V dataset.]{
		\begin{minipage}[b]{0.48\textwidth}
			\includegraphics[width=1\textwidth]{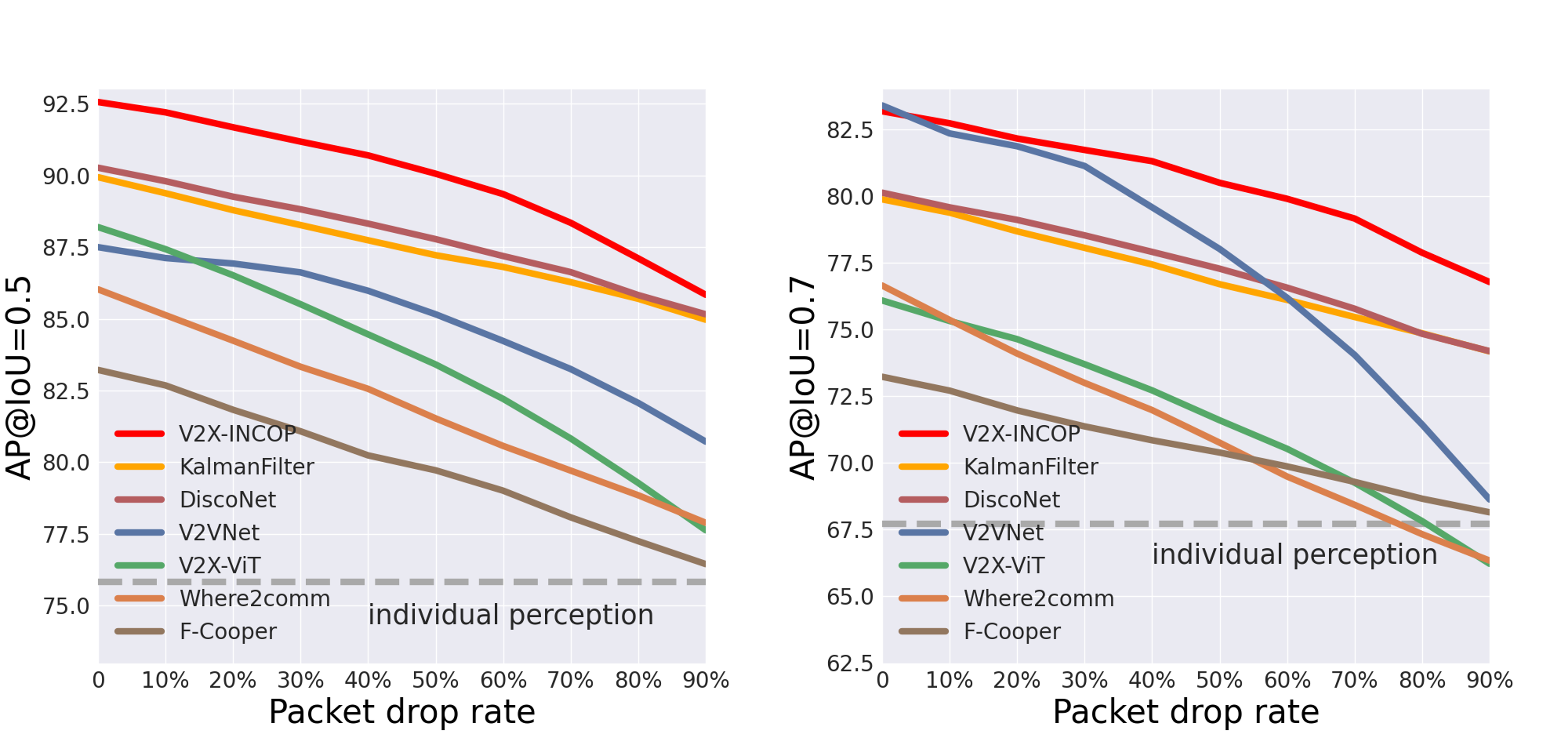}
		\end{minipage}
		% \label{fig:issue}
	}
    	\subfigure[Performance on Dair-V2X dataset.]{
    		\begin{minipage}[b]{0.48\textwidth}
   		 	\includegraphics[width=1\textwidth]{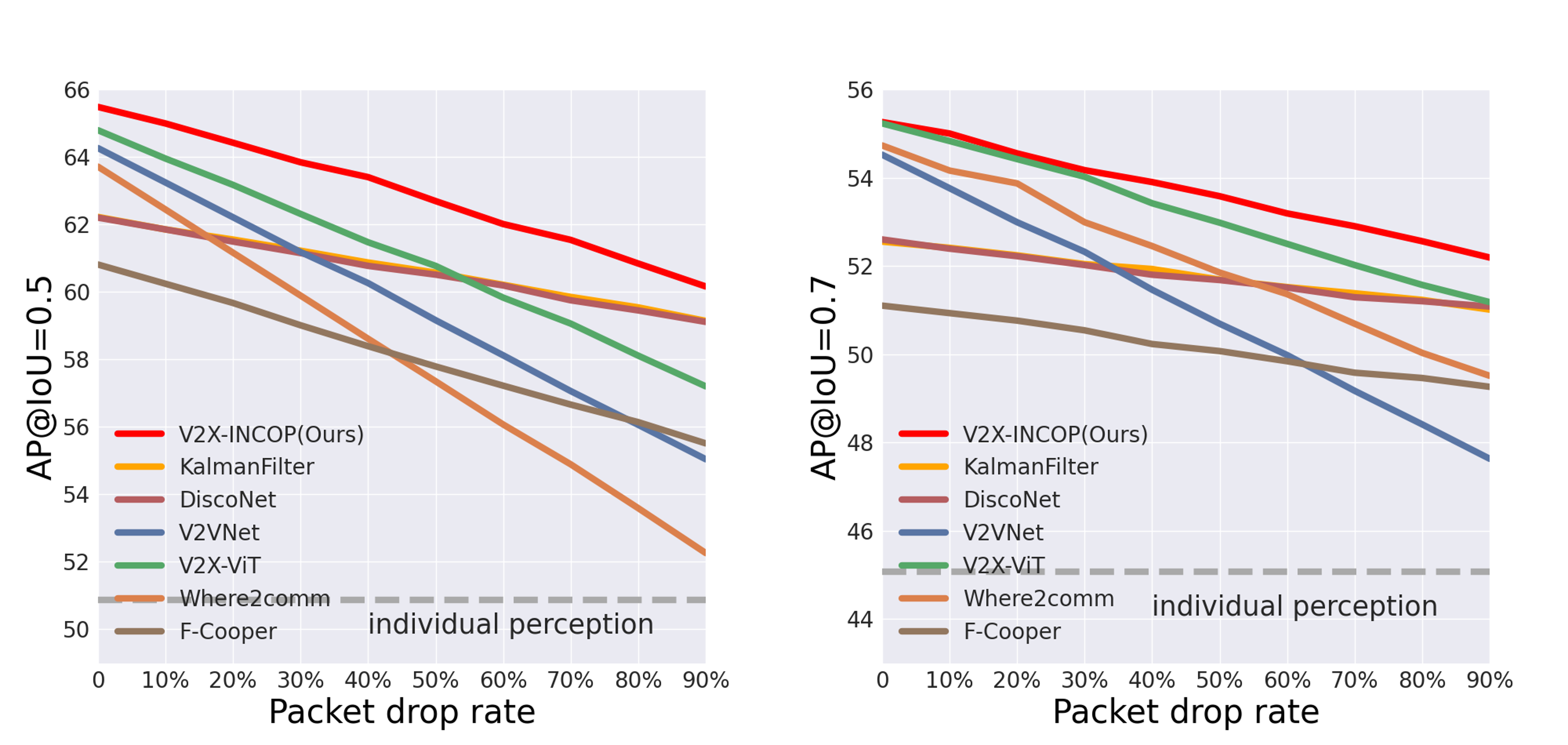}
    		\end{minipage}
		% \label{fig:performance degrade}
    	}
    \caption{
\textbf{Detection performance with different packet drop rates on OPV2V dataset and Dair-V2X dataset.} The horizontal axis represents the pack drop rate and the vertical axis represents the average precision of detection results. The results shows the proposed V2X-INCOP outperforms other methods at most packet drop rates.}
     
     \label{fig:opv2v_dair}
\end{figure*}

The implementation is introduced as follows.
For the backbone, we follow the setting of Pointpillars~\cite{lang2019pointpillars}, and set the width, length, and height of each voxel of the input to $(0.4m, 0.4m, 4m)$ for OPV2V and Dair-V2X and $(0.4m, 0.4m, 5m)$ for V2X-Sim. The number of channels of the encoded features is set to 256. 
For the SAFF model, the four $1 \times 1$ convolution layers are with 64, 32, 8, and 1 output channels, respectively.
For the prediction model, we set the number of the past timestep $k=3$. The kernel size of the two convolution layers is $3 \times 3$. The number of channels is first doubled after downsampling and then halved after upsampling.
For the knowledge distillation, we first train the teacher model without communication interruptions and then use the well-trained teacher model to guide the training of the prediction model.
For the curriculum learning strategy, we expand the range of the interruption probability by 20\% every 5 epochs in the training. For example, the interruption probability for each sample is randomly sampled from $[0,20\%]$ during the first 5 epochs and $[0,40\%]$ in the next 5 epochs, and so on.
In terms of coefficients in the loss function, we set $\alpha=1$, $\beta=2$ and $\gamma =10000$. We use Adam optimizer~\cite{Kingma2014AdamAM} and the learning rate is set to 0.002 initially. For a fair comparison, we implement all baselines and SOTA methods with Pointpillars, the same backbone as our proposed V2X-INCOP.  We train all the models using NVIDIA GeForce RTX 3090 GPU.

\subsection{Quantitative evaluation}

\textbf{Baselines and existing methods.}
We consider individual perception without cooperation as a baseline and compare our proposed method with existing state-of-the-art cooperative perception methods, including \textit{V2VNet}~\cite{wang2020v2vnet}, \textit{DiscoNet}~\cite{li2021learning}, \textit{F-Cooper}~\cite{chen2019f}, \textit{V2X-Vit}~\cite{xu2022v2xvit}, \textit{Where2comm}~\cite{Where2comm22}. 

In addition, we leverage Kalman filter to conduct missing information recovery in the detection output domain, referred to as \textit{Kalman filter}. Given the bounding boxes detected from the features fused with interruptions in the past $k$ timesteps, we employed a basic Kalman Filter for object tracking and the estimation of future states for each object. To facilitate object tracking, we applied the Hungarian algorithm to establish associations between predictions and measurements during the tracking process. Ultimately, we combined the predicted bounding boxes with the detection results based on the presently received features utilizing a straightforward selection approach.

\subsubsection{Benchmark comparison}

Figure~\ref{fig:v2x-sim} and
Figure~\ref{fig:opv2v_dair} show detection performance comparison with different PDRs on datasets V2X-Sim, OPV2V and Dair-V2X.
In V2X-Sim dataset, we evaluate performance in three scenarios: V2V, V2I, and V2X. In V2I and V2X scenarios, we evaluate the performance of connected vehicles, roadside units, and all cooperation nodes to study the performance of different kinds of cooperation nodes separately. 
Since OPV2V dataset only includes V2V scenarios, we evaluate the performance of the ego vehicle. Dair-V2X dataset collects data in a vehicle-infrastructure cooperation scenario where one vehicle cooperates with one roadside unit and only includes the V2I scenario, and we evaluate the performance of the ego vehicle following \cite{yu2022dair}.

From the results, we can see that: (a) Performance of all cooperative perception methods degrades as the PDR increases, which reflects the impacts of communication interruption. Some methods, like V2VNet and V2X-Vit, can achieve good performance with no interruption but performance drops fast as interruption happens. (b) The proposed V2X-INCOP outperforms all other methods at most PDRs, which suggests the robustness to interruption. Even with a high PDR, the performance of V2X-INCOP is still far better than individual perception, suggesting it can effectively alleviate the impacts of communication interruption. (c) On average of different PDRs, V2X-INCOP can bring performance gain over individual perception up to 13.9\%, 14.06\%, and 12.07\% on V2X-Sim, OPV2V, and Dair-V2X datasets, respectively, which can achieve a robust cooperative perception system.

\subsubsection{Robustness to pose noise}

A cooperative perception system needs precise pose information to fuse corresponding spatial information from cooperation nodes at different locations correctly. However, pose information estimated by the localization module is not perfect in practice, which often introduces pose noise and makes the cooperative perception system fail. To validate the robustness to pose noise of the proposed method, we test cooperative perception performance with different pose noise levels.
Following \cite{vadivelu2021learning} and \cite{lu2022robust}, we add Gaussian noise on 2D center and yaw angle of accurate 6DoF pose, $\mathcal{N}(0, \sigma_t), \mathcal{N}(0, \sigma_r)$, to simulate positional and heading error, respectively.

Figure~\ref{fig:ablation_noise} shows the performance of the proposed V2X-INCOP and other cooperative perception methods with different noise levels on OPV2V dataset, where the PDR is 50\%. We see that: the proposed V2X-INCOP shows only a slight performance decrease as the noise level increases and achieves the best performance at all different noise levels, reflecting its robustness to pose noise; while other methods like \textit{Late-fusion}, which share the detection results among cooperation nodes, are very sensitive to pose noise. 
Though \textit{Late-fusion} has good performance with no pose noise, the performance drops worse than individual perception even with a small pose error, which is impossible to apply to practice.

\begin{figure}
\centering
\includegraphics[width=0.8\linewidth]{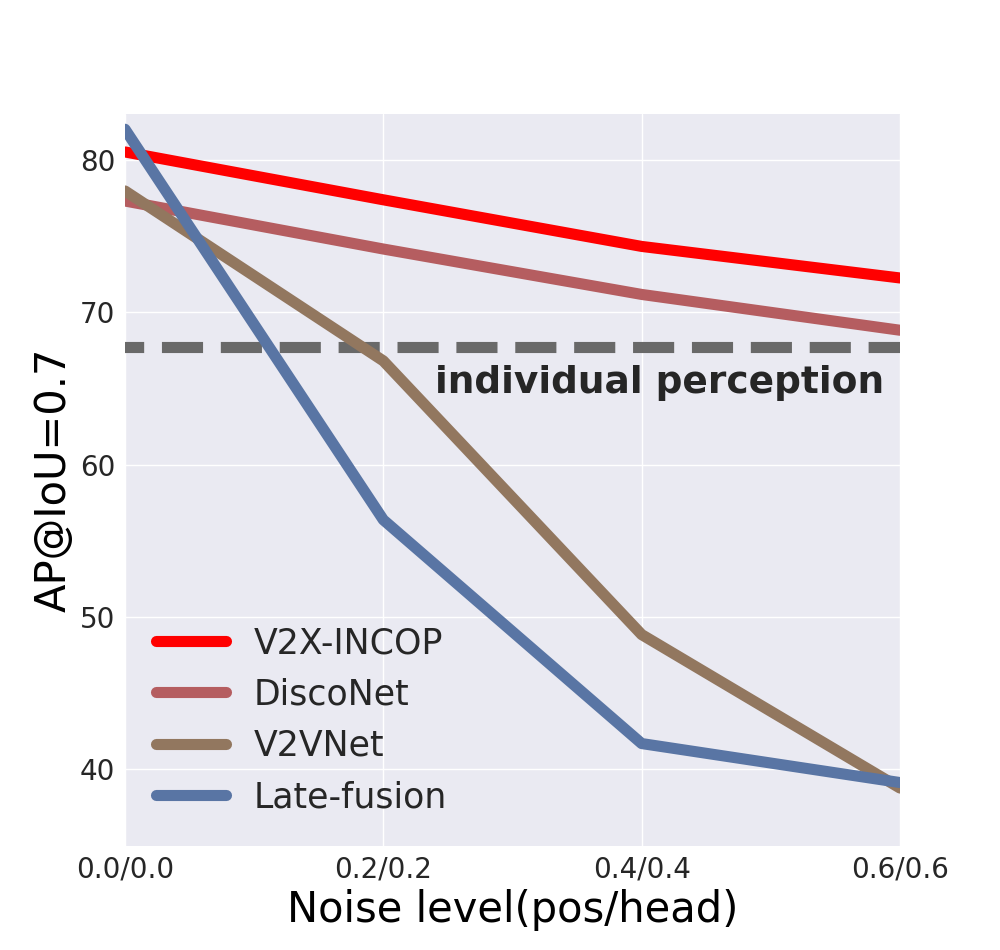}
\caption{\textbf{Detection performance with different pose noise level.} The horizontal axis represents the pose noise level and the vertical axis represents average precision of detection results. The pose noise level is measured by the deviation of positional and heading error, $\sigma_t/\sigma_r(m/\circ)$.}
\label{fig:ablation_noise}%
\end{figure}

\subsection{Qualitative evaluation}

\begin{figure*}
\centering
\includegraphics[width=\linewidth]{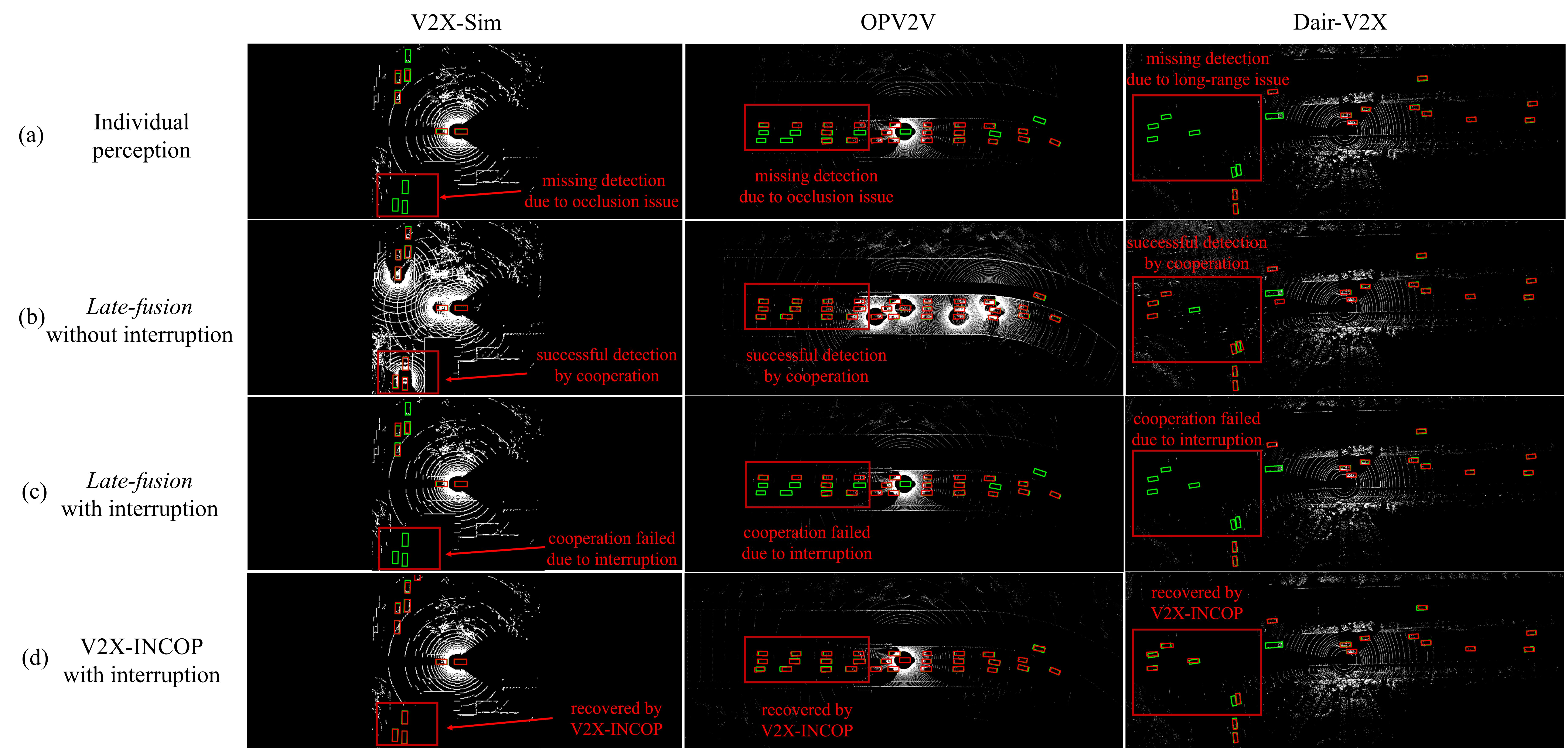}
\caption{\textbf{Visualization of detection results on datasets V2X-Sim, OPV2V and Dair-V2X.} Green/red boxes denote ground truth/detection results. The point clouds collected by both ego vehicles and cooperation vehicles are displayed in the second row for a better understanding of the effect of cooperative perception. With communication interruption, \textit{Late-fusion} fails to detect objects out of sight due to missing messages while the proposed V2X-INCOP recovers most of the lost boxes due to the interruption from historical information, making cooperative perception robust to interruption.}
\label{fig:vis}%
\end{figure*} 

Figure~\ref{fig:vis} shows the visualization of detection results to validate the effectiveness of the proposed method on datasets V2X-Sim, OPV2V, and Dair-V2X. Three columns represent three samples taken from the three datasets. The four rows visualize the output of four different detection methods, namely individual perception, \textit{Late-fusion} without communication interruption, \textit{Late-fusion} with communication interruption, and the proposed V2X-INCOP with communication interruption, respectively.  
It can be seen in Figure~\ref{fig:vis} that: 1) Individual perception may cause missing object detection due to some perception issues, such as long-range and occlusion. 2) Cooperative perception can detect objects in those occluded and long-range regions, which are invisible to ego vehicle. The point clouds collected by both ego vehicles and cooperation vehicles are displayed in the second row for a better understanding of the effect of cooperative perception.
3) In the case of communication interruption, the ego vehicle with \textit{Late-fusion} fails to detect objects out of sight. 
4) The proposed V2X-INCOP recovers most of the lost boxes due to the interruption from historical information, making cooperative perception robust to interruption.

\subsection{Ablation study}

\begin{figure}
\centering
\includegraphics[width=0.8\linewidth]{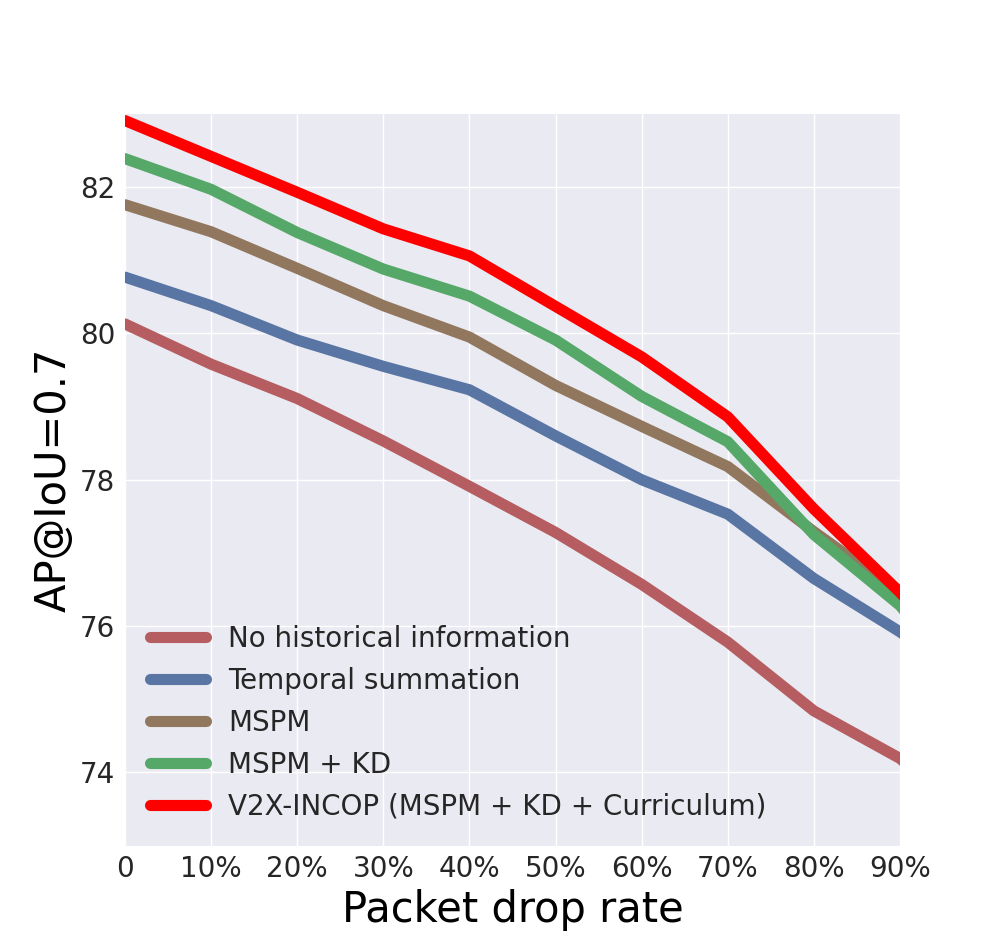}
\caption{\textbf{Ablation study of different components.} The horizontal axis represents the packet drop rate and the vertical axis represents the average precision of detection results. Curves in different colors represent different components combination. The results show detection performance increases and becomes more robust to interruption with more components in most communication conditions.}
\label{fig:ablation_component}%
\end{figure} 

We conduct ablation study to validate the effectiveness of each component. In the proposed V2X-INCOP, there are four novel designs: 1) introducing historical information to recover missing information; 2) multi-scale spatial-temporal prediction model (MSSTP); 3) supervising MSSTP with knowledge distillation; 4) stabilizing training with curriculum learning. To validate the effectiveness, we add each one of four designs step by step and test the perception performance of the following methods: 

\textit{No historical information} is regarded as a baseline, which does not introduce historical information and uses the spatial attention fusion model to fuse features of the current timestep.

\textit{Temporal summation} introduces historical information by summation of features of past timesteps, and fuse it with current features. 

\textit{MSSTP} means recovering the missing information from historical information by a multi-scale prediction model. 

\textit{MSSTP + KD} means providing explicit supervision to MSSTP with knowledge distillation. 

\textit{MSSTP + KD + Curriculum} is the complete V2X-INCOP, which adopts the curriculum learning training strategy.

Figure~\ref{fig:ablation_component} shows the detection performance of different methods with different PDRs on OPV2V dataset.
We can see that: 1) cooperative perception becomes robust to interruption by introducing historical information,  the information source from which we can recover the missing information; 2) the multi-scale prediction model can achieve better recovery of missing information than vanilla summation because it can capture the spatial-temporal features of the historical information for comprehensive prediction; 3) knowledge distillation can further improve perception performance because it can give direct supervision to the prediction model, but slightly with a high PDR due to possible hard training; 4) the complete V2X-INCOP achieves the best performance with the help of the curriculum learning strategy, which help stabilize the training stage and make the system attain better convergence.

\subsection{Effect of two hyperparameters}

\subsubsection{Effect of timesteps of historical features}

\begin{figure}
\centering
\includegraphics[width=0.8\linewidth]{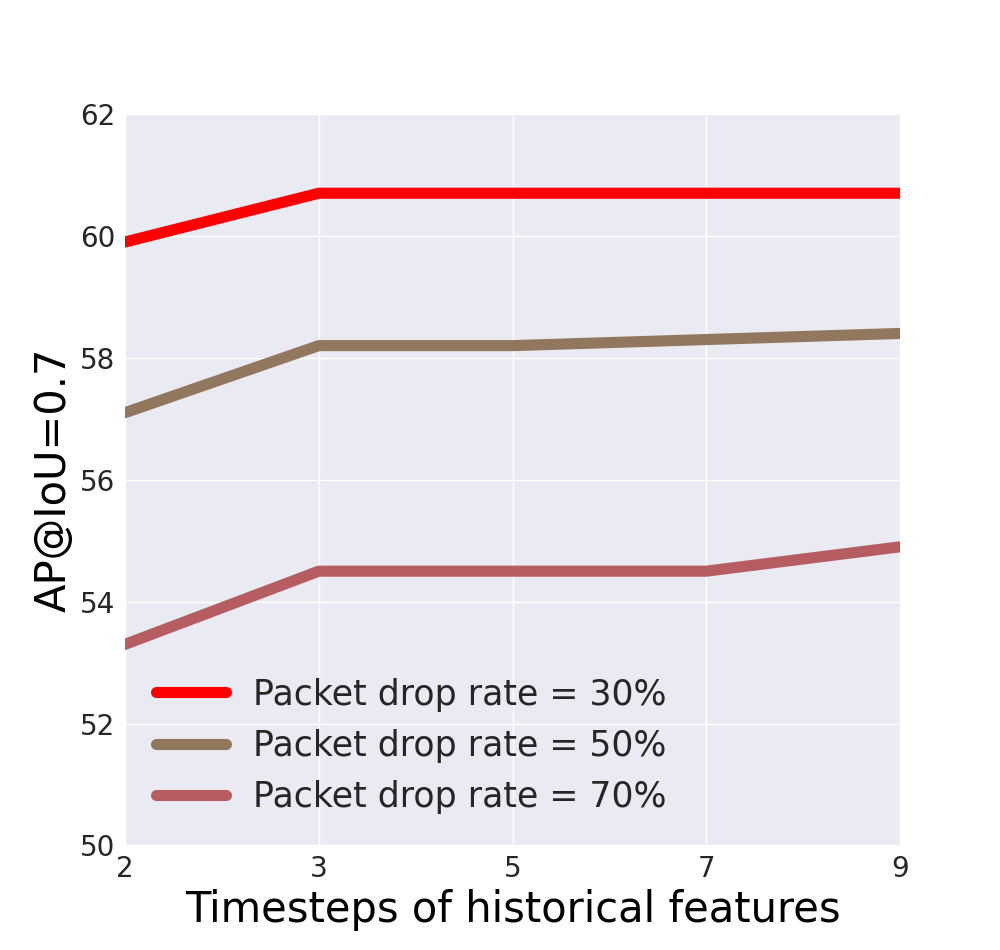}
\caption{\textbf{Detection performance with different timesteps of historical features.} The
horizontal axis represents the number of timesteps of historical features and the vertical axis
represents the average precision of detection results. Curves in different colors represent performance with different packet drop rates.}
\label{fig:ablation_frames}%
\end{figure} 

To explore perception performance affected by the number of previous timesteps, we test the performance of V2X-INCOP with different timesteps of historical features. Figure~\ref{fig:ablation_frames} shows detection performance with different timesteps of historical features on V2X-Sim dataset. We see that: (a) The detection performance increases with more timesteps of historical features, because more relevant information may be recovered from longer cooperation history, especially when the PDR is high.
(b) The gain becomes slight when the number of historical features is more than 3. This is because useful information gain is marginal and more noise may be introduced during prediction as more irrelevant information is taken into consideration.
In addition, more timesteps of historical features mean more computation and storage resources. For a better trade-off, we set the number of previous timesteps to 3 in the proposed V2X-INCOP.

\subsubsection{Effect of the number of cooperation nodes}

\begin{figure}
\centering
\includegraphics[width=0.8\linewidth]{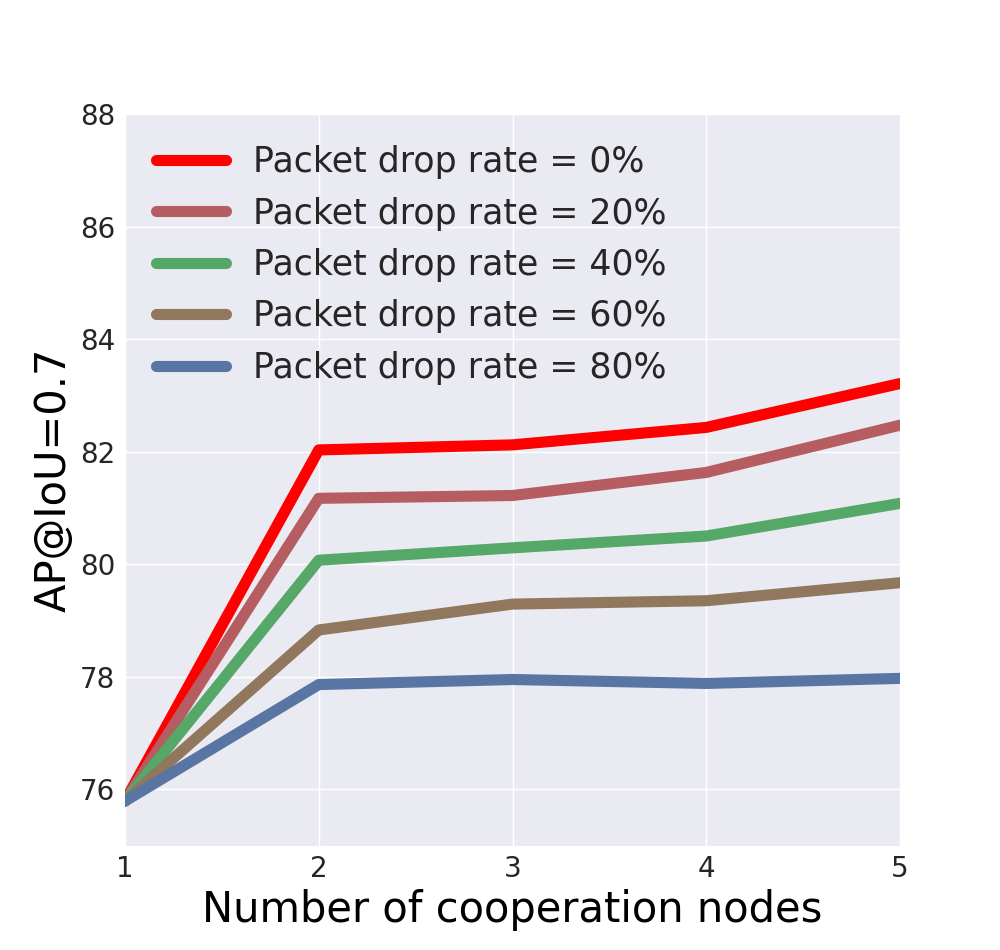}
\caption{\textbf{Detection performance with different number of cooperation nodes.} The
horizontal axis represents the number of cooperation agents and the vertical axis
represents the average precision of detection results. Curves in different colors represent performance with different packet drop rates.}
\label{fig:ablation_conum}%
\end{figure} 

We study perception performance with communication interruption affected by the number of cooperation nodes by gradually increasing the number of cooperation nodes on OPV2V dataset.
Figure~\ref{fig:ablation_conum} shows detection performance with different interruptions in the scenarios with 2$\sim$5 cooperation nodes. We see that: the detection performance increases as the ego node cooperates with more nodes and the gain brought by more cooperation nodes becomes slight as the PDR increases. This is because more useful information from other nodes is missing due to more frequent interruptions. However, with the proposed V2X-INCOP, the performance gain brought by cooperation still remains significant when the PDR is high.

\subsection{Discussion}

\subsubsection{Performing missing information recovery in the feature domain}

We perform missing information recovery in the feature domain for the following four key reasons. 
(a) Communication efficiency. When compared to sharing raw data, intermediate features are considerably easier to compress. By recovering missing information in the feature domain, we can maximize the efficiency of shared messages.
(b) Robustness to localization noise. Compared with conducting prediction and fusion in the output domain, it possesses more information to suppress and correct the noise.
(c) Simplified feature extraction. Recovery in the feature domain enables more straightforward feature extraction and usage since the recovery model is jointly trained in conjunction with the encoder and decoder. Additionally, it encourages the model to focus on predicting the most informative features. There is no requirement to predict precise and detailed scene information, which can be more challenging and less fault-tolerant.
(d) Flexibility across tasks and models. This approach offers flexibility for application across multiple tasks and models without necessitating changes to the design of the recovery module.

\subsubsection{Robustness and generalization}

We discuss the robustness and generalization of the proposed V2X-INCOP based on the experiment results from the following four aspects.

(a) Robustness to different communication conditions. From Figure~\ref{fig:v2x-sim}\&~\ref{fig:opv2v_dair}, we can see that the proposed V2X-INCOP maintains significant cooperative perception gain under different communication conditions($0\sim 90\%$ PDR), which indicates the proposed V2X-INCOP can adapt to varying degrees of communication conditions, from low to high PDRs.
(b) Robustness to pose error. From Figure~\ref{fig:ablation_noise}, we can see that the proposed V2X-INCOP shows only a slight performance decrease as the noise level increases and achieves the best performance at all different noise levels, reflecting that the proposed V2X-INCOP can be applied in scenarios in the presence of a certain level of positioning noise.
(c) Generalization to different scenarios. V2X-INCOP achieves the best performance on three cooperative perception datasets, which include different scenarios. V2X-Sim includes V2V and V2I scenarios in crowded downtown and suburb highways. OPV2V includes V2V scenarios with diverse city traffic configurations. Dair-V2X collects data from real-world vehicle-infrastructure operation scenarios. The results show that the proposed V2X-INCOP can be generalized to different scenarios.
(d) Generalization to different hyperparameters. Figure~\ref{fig:ablation_frames}\&\ref{fig:ablation_conum} show that V2X-INCOP achieves significant improvements when employing various numbers of timesteps of historical features and cooperation nodes, which indicates that the number of timesteps of historical features can be flexibly adjusted based on the actual deployment resources and environment and the proposed V2X-INCOP can be applied to scenarios with varying numbers of cooperation nodes.

\section{Conclusion}\label{sec:conclude}

This work proposes an INterruption-aware robust COoperative Perception (V2X-INCOP) system for V2X communication-aided autonomous driving, whose core idea is to recover missing information due to communication interruptions from historical information with a novel communication-adaptive multi-scale spatial-temporal prediction model. Comprehensive experiment results on three public cooperative perception datasets demonstrate that our method can bring significant benefits to alleviate the adverse effect caused by communication interruption. V2X-INCOP outperforms state-of-the-art cooperative perception methods and has a cooperative perception gain up to 14.06\%, 13.9\%, and 12.07\% over individual perception on average of different packet drop rates on OPV2V, V2X-Sim, and Dair-V2X datasets, respectively. This is the first work to address the communication interruption issue in cooperative perception systems to our best knowledge, which can boost the practical application of cooperative perception in the real world. In further research, more communication issues could be taken into consideration, such as communication delays and communication attacks, resulting in a more robust cooperative perception system.

\bibliographystyle{IEEEtran}
\bibliography{ref}

\end{document}